%% file: main.tex
\documentclass{article}

\usepackage[preprint,nonatbib]{neurips_2022}

\usepackage{amsmath}
\usepackage{amssymb}
\usepackage{mathtools}
\usepackage{amsthm}

\usepackage[utf8]{inputenc} %
\usepackage[T1]{fontenc}    %
\usepackage{hyperref}       %
\usepackage{url}            %
\usepackage{booktabs}       %
\usepackage{amsfonts}       %
\usepackage{nicefrac}       %
\usepackage{microtype}      %
\usepackage[dvipsnames]{xcolor}
\hypersetup{
    colorlinks,
    linkcolor={red!50!black},
    citecolor={blue!50!black},
    urlcolor={blue!80!black}
}

\usepackage{thm-restate}
\usepackage{graphicx}
\usepackage{relsize}
\usepackage{multirow}
\usepackage{textcomp}
\usepackage[ruled]{algorithm2e}

\definecolor{tab:blue}{HTML}{1F77B4}
\definecolor{tab:orange}{HTML}{FF7F0E}
\definecolor{tab:green}{HTML}{2CA02C}
\definecolor{tab:red}{HTML}{D62728}
\definecolor{tab:purple}{HTML}{9467BD}
\definecolor{tab:brown}{HTML}{8C564B}
\definecolor{tab:pink}{HTML}{E377C2}
\definecolor{tab:gray}{HTML}{7F7F7F}
\definecolor{tab:olive}{HTML}{BCBD22}
\definecolor{tab:cyan}{HTML}{17BECF}

\theoremstyle{plain}
\newtheorem{theorem}{Theorem}[section]

\theoremstyle{definition}

\theoremstyle{remark}
\newtheorem{remark}[theorem]{Remark}

\usepackage[ruled]{algorithm2e}

\input{math_commands}

\newcommand{\ptheta}{{p_\vtheta}}
\newcommand{\stheta}{\vs_\vtheta^p}

\newcommand{\ntheta}{\nabla_\vtheta}

\newcommand{\qphi}{{q_\vphi}}

\newcommand{\gtheta}{\vg_\vtheta}

\newcommand{\ppost}{{\ptheta(\vz|\vx)}}

\newcommand{\numberthis}{\addtocounter{equation}{1}\tag{\theequation}}

\newcommand{\colorLK}[1]{{\color{Cerulean}\gL}_{{\color{cyan}\mathrm{KL}}#1}}
\newcommand{\colorLtwo}[1]{{\color{blue}\gL}_{{\color{blue}\mathrm{2}}#1}}
\newcommand{\colorhK}{{\color{magenta}h_\mathrm{K}}}
\newcommand{\colorhF}{{\color{green}h_\mathrm{F}}}
\newcommand{\Esub}[1]{{\E_{#1}}\!}
\newcommand{\Esubscore}[2]{{\E_{#1}}\!\left[\left\| #2 \right\|_2^2\right]}
\newcommand{\Jqtp}{J^\pi_{q,\vtheta}(\vx)}

\usepackage[natbib=true,style=numeric,maxcitenames=1,uniquename=false,uniquelist=false]{biblatex}
\usepackage[capitalize,noabbrev]{cleveref}

\usepackage[textsize=tiny]{todonotes}

\def\mytitle{On the failure of variational score matching \\ for VAE models}
\title{\mytitle}

\title{\mytitle}

\author{%
  Li Kevin Wenliang\thanks{Partially carried out during an internship 
  at Amazon Web Services Shanghai Lablet.} \\
  Gatsby Computational Neuroscience Unit\\
  University College London\\
  \texttt{kevinli@gatsby.ucl.ac.uk} \\
}

\begin{document}

\maketitle

\begin{abstract}

Score matching (SM) is a convenient method for training flexible probabilistic models, 
which is often preferred over the traditional maximum-likelihood (ML) approach.
However, these models are less interpretable than normalized models; 
as such, training robustness is in general difficult to assess.
We present a critical study of existing variational SM objectives, showing 
catastrophic failure on a wide range of datasets and network architectures.
Our theoretical insights on the objectives emerge directly from their 
equivalent \emph{autoencoding losses} 
when optimizing variational autoencoder (VAE) models.
First, we show that in the Fisher autoencoder, SM produces far worse models than maximum-likelihood, 
and approximate inference 
by Fisher divergence can lead to low-density local optima.
However, with important modifications,
this objective reduces to a regularized autoencoding loss that 
resembles the evidence lower bound (ELBO).
This analysis predicts that the modified SM algorithm should behave very similarly 
to ELBO on Gaussian VAEs.
We then review two other FD-based objectives from the literature and 
show that they reduce to uninterpretable autoencoding losses, likely leading to poor performance.
The experiments verify our theoretical predictions and suggest that only ELBO and the baseline
objective robustly produce expected results, while previously proposed SM methods do not.

\end{abstract}

\section{Introduction}

Finding robust algorithms for training expressive latent variable models remains 
at the heart of unsupervised learning research. 
Latent variable models describe the data distribution by mapping a prior distribution 
of latent variables through a likelihood. 
A well-established approach for training such models is maximum likelihood (ML) that 
minimizes the Kullback-Leibler divergence (KLD) between the data and model distributions. 
Exact ML is usually intractable in practice; 
instead, one optimizes a lower bound of the log-likelihood (or upper bound on the KLD) 
defined through a variational distribution, known as the evidence lower bound (ELBO) \citep{Kingma2014, Rezende2014}.
These approaches have produced robust results across task, datasets and model architectures \citep{kingma2019introduction}.
Score matching offers another approach to training statistical models \citep{hyv}; 
it minimizes the Fisher divergence (FD), a convenient objective for potentially unnormalized models.
When latent variables are present, a situation we focus on here, 
the FD can be approximated \citep[e.g.\ ][]{swersky2011autoencoders, vertes2016learning, bao2020bi, fae}.

Practitioners, who want to model their datasets under budget,
see these two alternatives and naturally ask: which one 
is more robust so that the results are generally good without expensive hyperparameter search?
KLD and FD are connected through 
differential relationships \citep{stam1959some,guo2009relative,liu2016stein}, but these 
results do not predict the robustness of using these divergences for training.
Some previous work examined the difference between ML and 
SM in fully observed models \citep{wenliang2019learning,arbel2018kernel, song2020sliced} 
and latent variable models \citep{bao2020bi}; however, comparisons
that control for the model class are missing. 
In addition, existing SM objectives that give empirical success appear similar to each other
\citep{fae, bao2020bi} but their differences remain unclear. 
Addressing these issues gives better understand of these methods, 
and avoid unexpected failures and expensive hyperparameter search. 
Further, a well-controlled comparison can reveal whether the improvements 
seen previously should be attributed to training objectives or other details.

Here, we conduct such a study of variational SM objectives using variational autoencoding models (VAEs) 
to unveil their differences and connection.
Importantly, the joint distribution of a VAE model is normalized and thus can be trained conveniently via ML and SM;  
as such, VAE models provide an intuitive setup for theoretical analysis and well-controlled experiments:
a robust ML or SM algorithm must be capable of training a VAE on a dataset. In particular, we focus 
on Gaussian VAEs on which important theoretical insights have been derived 
and practical successes seen \citep[e.g.\ ][]{dai2018diagnosing,lucas2019don,dai2021,rybkin2021simple} .

In \cref{sec:KL_bounds}, we review the traditional variational ML approach and its objective (ELBO).
\textbf{Our main contributions} consist of the analyses and illustrations 
in \cref{sec:variational_SM,sec:related}, and the 
experiments in \cref{sec:exp}. 
Specifically,
\cref{sec:variational_SM} studies a recently proposed variational SM objective known as the Fisher autoencoder \citep{fae}.
We show that this objective can perform poorly in a counterintuitive way.
We then, for the purpose of comparison, introduce a benchmark SM objective that is very similar to ELBO 
in their equivalent autoencoding losses and is thus expected to work as well as the latter.
Unfortunately, two other existing SM objectives
\citep{swersky2011autoencoders, bao2020bi} 
do not admit any reasonable autoencoding loss \cref{sec:related}.
Experiments (\cref{sec:exp}) demonstrate that the benchmark variational SM objective 
gives comparable performance to ELBO on several metrics, 
while the other variational objectives often failed. 
A concluding summary and limitations of this work are discussed in \cref{sec:discuss}.

\section{Background: variational autoencoders and maximum likelihood}\label{sec:KL_bounds}

We analyze training objectives on VAEs in which the generative model is expressed as
$\ptheta(\vz,\vx)=p(\vz)\ptheta(\vx|\vz),$
where $\vz\in\sR^{d_z}$ is latent, $\vx\in\sR^{d_x}$ is observed, 
and $\vtheta$ is the vector of parameters. 
The prior is fixed for the purpose of this study.
The parameters $\vtheta$ of this model is trained using i.i.d.\ samples of a dataset $\gD$ so that the marginal 
$\ptheta(\vx)=\int p(\vz)\ptheta(\vx|\vz)\ud\vz$ is close to the underlying unknown data density $\pi(\vx)$. 
To find $\vtheta$, maximum-likelihood (ML) optimizes the marginal KLD
$
\KL[\pi(\vx)\|\ptheta(\vx)]:=\Esub{\pi}\left[\log \pi(\vx) - \log \ptheta(\vx)\right],
$
which is intractable for most models used in practice. 
A more practical approach to ML is to optimize ELBO: 
\begin{align}\label{eq:elbo}
        \gF_{q,\vtheta}(\vx):=\E_{q}\left[ \log \ptheta(\vx|\vz) \right] - \KL\left[q(\vz|\vx)\| p(\vz) \right],
\end{align}
where $q(\vz|\vx)$ is a variational distribution in some prespecified distributional family $\gQ$. 
It is usually parametrized by $\vphi$ which is independent of $\vtheta$.
It can be shown that maximizing the expected ELBO amounts to minimizing the joint KLD
\begin{align*}
\KL[q(\vz|\vx)\pi(\vx)\|\ptheta(\vz,\vx)]
    = \E_{\pi}\!\left[ \KL[q(\vz|\vx)\|\ptheta(\vz|\vx)]\right] + \KL[\pi(\vx)\|\ptheta(\vx))].\numberthis\label{eq:KL_bounds}
\end{align*}
The joint KLD is attractive for learning $\vtheta$ because it upper bounds the 
marginal KLD, and the variational posterior $q$ can be optimized independently of $\vtheta$.
In this work, we focus on the Gaussian VAE where the relevant densities are given by
\begin{gather*}
    p(\vz)=\gN(0,\mI),\quad
    \ptheta(\vx|\vz)=\gN(\vx;\vg_\vtheta(\vz), \gamma\mI),\numberthis\label{eq:vae_densities}\quad
    q(\vz|\vx)=\gN(\vz;\vmu(\vx),\mSigma(\vx))).
\end{gather*}
Here, the functions $\vg_\vtheta$, $\vmu$ and $\mSigma$ specify the distribution parameters, 
and $\gamma>0$ is a scalar variance of the likelihood.
In later sections, we sometimes relax the assumption that $p(\vx)$ is Gaussian.
The Gaussian VAE is not only widely used in practice for modelling complex continuous data distributions 
but also intuitive for deriving interesting properties. 
For example, it can be shown \citep{dai2018connections, dai2021} 
that in the limit of small posterior variance,
the ELBO on the Gaussian VAE reduces to the following 
autoencoding objective at its optimum
\begin{gather*}
    \min_{\vtheta,q} {\colorhK}\!\left(\frac{1}{N}\sum_{n=1}^N\colorLK{,q,\vtheta}(\vx_n)\right) 
     + \frac{1}{N} \sum_{n=1}^N \|\vmu(\vx_n)\|_2^2\numberthis\label{eq:vae_ae}, \\
    {\colorLK{,q,\vtheta}}(\vx):= 
        \left\| \vx - \E_{q}\!\left[\vg_\vtheta(\vz))\right] \right\|_2^2 
            + \sum_{j=1}^{d_x}\sV_{q}\!\left[\gtheta(\vz)_j\right],
\end{gather*}
where ${\colorhK}(y):=d_x\log (y/d_x)/2$ is an increasing concave function approaching $-\infty$ as $y\to0_+$.
Importantly, \citet{dai2021} further showed that 
the unbounded gradient around zero in $h$ is necessary for
learning a form of \emph{optimal sparse representation} of the data. 
Such a property is valuable to many downstream applications that 
rely on efficient lower-dimensional representations. We will see how the autoencoding loss \eqref{eq:vae_ae} 
is related to objectives of variational score matching in Gaussian VAEs.

\section{Variational score matching}\label{sec:variational_SM}

As an alternative to ML, SM minimizes the marginal FD between the data and model distributions
\begin{equation}\label{eq:marginal_fd}
\FD[\pi(\vx)\|\ptheta(\vx)]:=\frac{1}{2}\Esub{\pi}\left[ \left\| \nx\log\pi(\vx)-\nx\log\ptheta(\vx)\right\|_2^2 \right],
\end{equation}
where $\nx$ is the Jacobian w.r.t.\ $\vx$. Practical algorithms 
for fully observed models were initially proposed by \citet{hyv}. 
Here, we consider SM algorithms for VAE models. 
For brevity, we define the following shorthands
$\stheta(\vx|\vz)\!:=\!\nx\log\ptheta(\vx|\vz)$ and $\vs^q(\vz|\vx)\!:=\!\nx\log q(\vz|\vx)$.

\subsection{Revisiting Fisher autoencoder}\label{sec:JFD_learning}

As discussed in \cref{sec:KL_bounds}, the joint KLD (JKLD) provides
a convenient objective for approximate ML.
\citet{fae} applied this intuition to SM and proposed the Fisher autoencoder (FAE) which 
optimizes the joint FD (JFD)
\begin{gather*}
\FD[q(\vz|\vx)\pi(\vx)\|\ptheta(\vz,\vx)]=\Esub{\pi}\left\{\FD[q(\vz|\vx)\|\ptheta(\vz|\vx)]+ \Jqtp\right\} \numberthis\label{eq:FD_bounds},\\
\Jqtp := \tfrac{1}{2}\E_{q}\!\left\|\nx\log\pi(\vx) 
                                            \!+\! \vs^q(\vz|\vx) \!-\! \stheta(\vx|\vz) \right\|_2^2
    =\tfrac{1}{2}\E_{q}\!\left\|\nx\!\log\tfrac{\pi(\vx)}{\ptheta(\vx)} 
        \!+\! \nx\!\log\tfrac{q(\vz|\vx)}{\ptheta(\vz|\vx)}\right\|_2^2\numberthis. \label{eq:FD_bounds_two}
\end{gather*}
A derivation is given in \cref{sec:FD_two_terms_deriv}. The two terms in \eqref{eq:FD_bounds} arise from 
the partial derivatives w.r.t.\  $\vz$ and $\vx$, respectively.
For a fixed $\vtheta$, if $q(\vz|\vx)=\ptheta(\vz|\vx)$ for all $\vx$ and $\vz$, 
then $\E_{\pi}[\Jqtp] = \FD[\pi(\vx)\|\ptheta(\vx)]$.
\citet{fae} further showed that $\E_\pi[\Jqtp]$ is 
equal (up to a $\pi$-dependent constant
that can be ignored for optimization) to $\E_\pi[M_{1,q,\vtheta}(\vx)]$, where
\begin{equation}\label{eq:M1}
    M_{1,q,\vtheta}(\vx)
        = \Eq\left[\tfrac{1}{2}\| \stheta(\vx|\vz) \|_2^2 + \nabla_\vx\cdot\stheta(\vx|\vz)\right]
          + \tfrac{1}{2}\Eq\left[\| \vs^q(\vz|\vx) \|_2^2\right]
\end{equation}
is a practical objective for SM. %
However, unlike the KLD case \eqref{eq:KL_bounds}, the JFD is \emph{not} an upper bound on 
the marginal FD. 
This means that a good fit over JFD may not imply a good fit over the target marginal FD.
A perfect match in the joint distributions implies a perfect match on their marginals. But when the joint 
distributions do not match \emph{exactly},
there is generally no guarantee that the marginals will be close, unless there is a bounding relationship
as in the KLD case.

\begin{figure}
    \centering
    \includegraphics[width=0.7\columnwidth, trim=0 0.5cm 0 1cm]{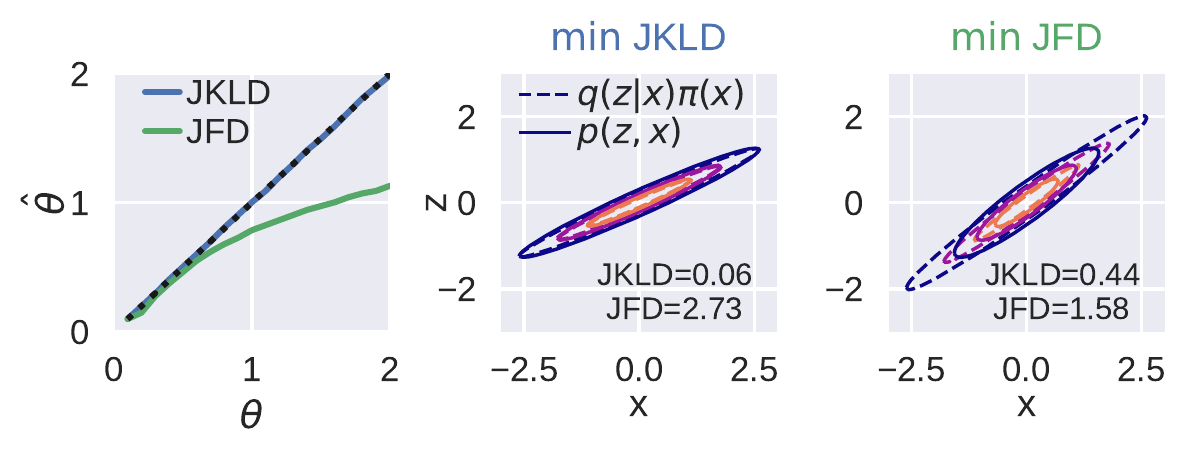}
    \caption{Parameter recovery by minimizing joint FD or joint KLD and the solutions.
    The bias in JFD minimization is substantial (left). 
    For $\theta=2$,
    minimizing the JKLD 
    gave a more sensible solution (middle) than minimizing the JFD (right). 
    }
    \label{fig:gaus_joint}
\end{figure}

For example, consider a simple VAE defined by a prior 
$p(z)=\gN(0,1)$, likelihood $p_\theta(x|z)=\gN(\theta x, \gamma)$, 
and a parametric variational posterior 
$q_\phi(z|x)=\gN(\phi x, \alpha v^*)$,
where $v^*:=(1+\theta^2/\gamma)^{-1}$
is the exact posterior variance for the generative model, $\gamma=0.5$ is fixed,
and the factor $\alpha$ is used to impose posterior mismatch. 
Suppose we perform parameter recovery: Let the data distribution $\pi$ 
be $p_\theta$ for a ground truth $\theta=\theta^*$, and we wish to 
recover $\theta^*$ by $\min_{\phi,\theta}\FD[q_\phi(\vz|\vx)\pi(\vx)\|p_\theta(\vz,\vx)]$. 
When $\alpha=1$, the posterior is exact, and thus any setting of $\theta^*$ can be recovered. 
We focus on the case where $\alpha=0.6$ and show that $\hat\theta$ estimated  
by JFD minimization can be far away from $\vtheta^*$ (\cref{fig:gaus_joint}, left) .
On the contrary, minimizing the JKLD produced robust estimates. 
To further illustrate this problem, we plot $q(\vz|\vx)\pi(\vx)$ and $p_{\hat\theta}(\vz,\vx)$ found by 
minimizing JKLD and JFD 
in \cref{fig:gaus_joint} when the true $\theta=2.0$. These joints are close in JFD but not at all 
in JKLD or visual judgment. 
Therefore, minimizing JFD jointly over $\vtheta$ and $q$ \citep{fae} may be 
undesirable when $q$ is approximate. 

\subsection{Improved variational SM for semi-Gaussian VAEs}\label{sec:improve_fae}
We took a closer look at the loss $M_1$ proposed by \citet{fae} 
and discovered that a simple modification can lead to an autoencoding loss 
similar to \eqref{eq:vae_ae} for a wider class of VAEs. 
For a \emph{semi-Gaussian VAE} in which only the likelihood $\ptheta(\vx|\vz)=\gN(\vx;\vg_\vtheta(\vz), \gamma\mI)$ is Gaussian but $p(\vz)$ and $q(\vz|\vx)$ are unspecified, we have the following proposition, derived in \cref{sec:ae_objectives}.
\begin{restatable}{proposition}{lone}
\label{thm:l1}
For a fixed q, the optimal semi-Gaussian VAE \eqref{eq:vae_densities} 
that minimizes $\E_\gD\left[M_{1,\phi, \theta}(\vx) \right]$ 
solves the following autoencoding loss.
\begin{gather}\label{eq:M1_ae}
    \min_{\vtheta} \colorhF\!\left(\frac{1}{N}\sum_{n=1}^N\colorLK{,q,\vtheta}(\vx_n)\right), \quad
    \colorLK{,q,\vtheta}(\vx) = 
        \left\| \vx - \E_{q}\!\left[\vg_\vtheta(\vz))\right] \right\|_2^2 
            + \sum_{j=1}^{d_z}\sV_{q}\!\left[\gtheta(\vz)_j\right],
\end{gather}
where $\colorhF(y)= -d_x^2/(2y)$, and $\colorLK{}$ is repeated from \eqref{eq:vae_ae}.
\end{restatable}
\begin{remark}\label{thm:m1_like_elbo}
The objective in \eqref{eq:M1_ae} is almost identical to the first term in \eqref{eq:vae_ae}
derived from ELBO except for the utility function. 
Like $\colorhK$ in \eqref{eq:vae_ae}, the function $\colorhF$ in \eqref{eq:M1_ae} is concave and 
has an unbounded 
gradient around zero, which is necessary for a sparse representation \citep{dai2021}.
\end{remark}

\cref{thm:l1} suggests that $M_1$ should in fact be used as an 
objective for optimizing $\vtheta$ (learning) only, effectively removing the expectation
under $s^q$ in \eqref{eq:M1}. 
On the other hand, the objective for optimizing $q$ (inference) is unspecified.
The rightmost equality in \eqref{eq:FD_bounds_two} indicates that
if $\nx\log[q(\vz|\vx)/\ptheta(\vz|\vx)]$ is close to zero, 
then $\Jqtp$ roughly equals the marginal FD.
Therefore, a gradient step on $\vtheta$ should not proceed until $q$ closely matches $\ptheta(\vz|\vx)$. 
This overall procedure does not optimize the joint FD, as in FAE \citep{fae}, or upper-bounds any marginal divergence; 
this is unlike variational ML which optimizes the joint KLD over $q$ and $\vtheta$ and bounds the marginal KLD
from above. 
In practice, the variational $q$ can be updated by FD or KLD, as discussed in the next subsection. 

We stress the algorithm described above \textbf{does not optimize the joint FD as in FAE \citep{fae}, 
nor does it perform coordinate descent between $\theta$ and $q$ on the original JFD}. 
In this procedure, $M_1$ is used to optimize for $\theta$ only, which effectively removes 
the second summand in the FAE objective \eqref{eq:M1}; this term is also not involved in optimizing $q$. 
Further, $\theta$ is not optimized through the posterior FD in \eqref{eq:FD_bounds}. 
Our modification thus deviates from FAE in nontrivial ways.

\subsection{Fisher divergence for inference}\label{sec:FD_inference}

Training an encoder by minimizing the KLD is a popular approach for many VAE models and even jointly
unnormalized models \citep{bao2020bi}.
However, for real $\vz$, the FAE objective \eqref{eq:FD_bounds} suggests FD as an objective for inference.
While both divergences can be estimated and optimized 
easily for the Gaussian VAE model, it is unclear which objective should be preferred for inference.
Here, we discuss properties of the solution and optimization when $q$ is optimized for FD.

\subsubsection{Optimal posterior in Gaussian VAEs}\label{sec:FD_inference_opt}
For a Gaussian VAE,
in situations where the data have low noise (e.g. natural images), 
the optimal posterior usually has small variance. In this case, 
as explained in \cref{sec:nonzerofd_proof},
the posterior FD approximately reduces to 
\begin{equation}\label{eq:post_fd}
\FD[q(\vz|\vx)\|\ptheta(\vz|\vx)] \approx \tfrac{1}{2}\|\vmu^*(\vx)\|_2^2,
\end{equation}
where the optimal posterior precision of $q$ is 
$
\mLambda^* \approx \mI + \tfrac{1}{\gamma}(\nz\vg_\vtheta(\vz)\nz\vg_\vtheta(\vz)^\intercal)|_{\vmu^*(\vx)}.
$
As such, when the Gaussian VAE model is well trained and the posterior variance is small, 
FD as the inference objective regularizes the posterior mean 
in a way similar to the KLD \eqref{eq:vae_ae}. 
Further, the optimal $\mLambda^*$ in FD coincides with that in KLD as derived by 
\citet[Equation 83]{dai2018diagnosing}.
Thus, using FD for inference can produce the same desirable effects arising from this $\mLambda^*$ discussed by the
authors.
However, \Cref{eq:post_fd} also reveals that the optimal Gaussian $q$ 
still incurs nonzero FD.
The joint and marginal FDs in \eqref{eq:FD_bounds} are then separated by the 
nonzero posterior FD \cref{eq:post_fd}. Therefore, we expected that FD is not an 
ideal inference objective for learning a Gaussian VAE by JFD or $M_1$.

\subsubsection{Local optima in FD optimization}\label{sec:FD_inference_toy}
\begin{figure}[t]
    \centering
    \includegraphics[width=0.49\columnwidth]{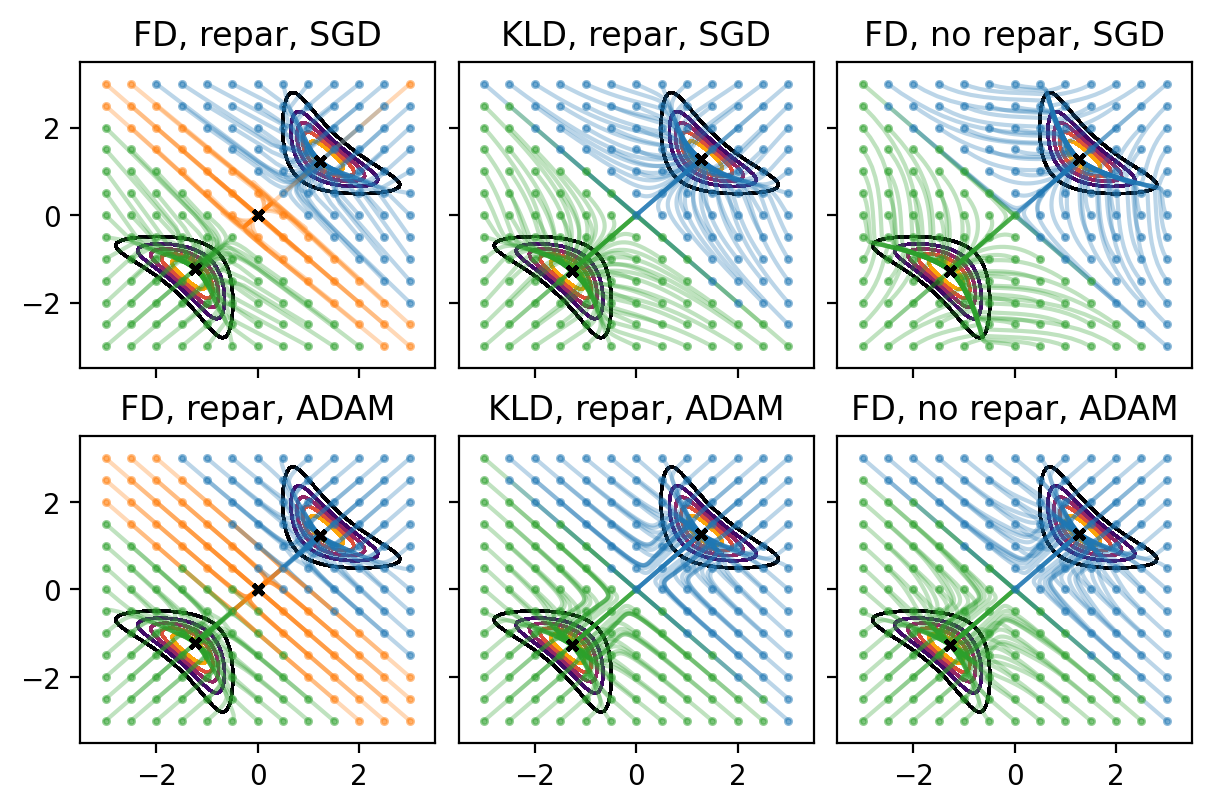}
    \includegraphics[width=0.49\columnwidth]{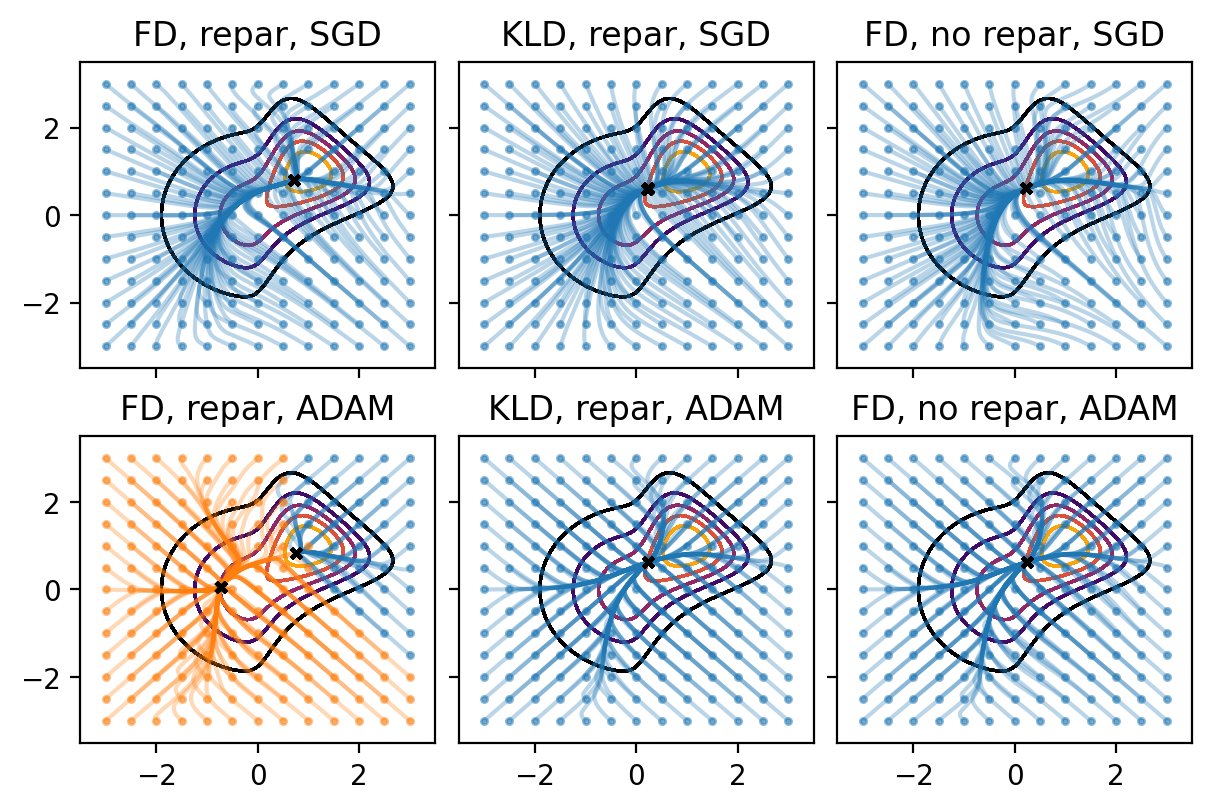}
    \caption{Optimizing a Gaussian $q$ under a Gaussian prior and the 
    simple likelihoods \eqref{eq:toy_liks} by $\KL$ and $\FD$ with or without 
    reparametrization. Panels on the left is for $p_\mathrm{I}$. 
        The heat contours show the true posterior distributions.
        The regular grid of dots show the initial means of the Gaussian $q$;
        the curved lines show the trajectories of the means during optimization that end at black crosses;
        The colors blue, green and orange are determined by the location of the mean 
        on convergence. 
    }
    \label{fig:opt_traces}
\end{figure}
Having characterized the solution of FD inference on a Gaussian VAE, we now turn to optimization.
\citet{fae} proposed to optimize a parametric $\qphi$ using 
the unbiased reparametrization gradient.
However, we were unable to produce any reasonable results with this approach, 
which prompted us to investigate optimization on simpler problems. 
Consider a generative model with a standard Gaussian prior and a likelihood in 
\begin{equation}\label{eq:toy_liks}
p_\mathrm{I}(x=2|z_1,z_2) = \gN(x=2;z_1z_2, 0.5^2),\quad
p_\mathrm{II}(x=1|z_1, z_2)=\gN(x=1;z_1\mathrm{relu}(z_2), 1.0^2).
\end{equation}
The posteriors $\ppost$ induced by these likelihoods as shown in \cref{fig:opt_traces} (contours).
To approximate these posteriors, 
we chose a variational family $\gQ$ as factorized Gaussians 
and optimized its mean and variance on $\FD[\qphi(\vz|\vx)\|\ptheta(\vz|\vx)]$ 
by either SGD or Adam \citep{kingma2015adam}, using reparametrization gradients. 
We visualize in \cref{fig:opt_traces} (``FD, repar'') the trajectories 
of its mean when initialized at different points in the latent space.
Surprisingly, for likelihood $p_\mathrm{I}$, there is a low-density local optimum 
at the origin that attracts variational posteriors with means initialized from the orange dots. 
Using Adam exacerbated this problem by enlarging the basin of attraction.
A similar issue also arises in the optimization of the Stein discrepancy \citep{korba2021kernel}. 
In contrast, minimizing the KLD did not produce such a solution (``KLD, repar''). 
For the likelihood $p_\mathrm{II}$, Adam also produced a local optima far away from the bulk of the 
posterior mass. SGD found a unique optimum regardless of initialization, but the solution 
was visibly different from that of KLD.

We hypothesize that the inappropriate convergence to low-density local optima is due to the 
gradient contribution of reparametrization: the 
FD can be lowered when $q$ is driven to concentrate around a small region 
where its \emph{shape} (but not its density) 
roughly matches the posterior's (i.e.\ $|\nx\log\ppost - \nz\log q(\vz|\vx)|$ 
is small).
We test this hypothesis by optimizing $q$ without reparametrization.
In \cref{sec:fd_grad_like_kl}, we show that the 
FD gradient without reparametrization is more similar to the KLD gradient 
than reparametrized gradient for a simple problem.
Indeed, for the toy problems in \cref{fig:opt_traces}, following this 
gradient (``FD, no repar'') resulted in more similar dynamics and 
final solutions to KLD inference.
More importantly, the posteriors no longer converge to the bad local optima.
In our experiments on image data, we could only obtain a reasonable fit by 
using unparametrized gradients when FD was chosen as the objective.
Although some advantage arises now that inference may not 
require reparametrized samples (\cref{sec:add_exp}), 
this biased gradient can be problematic, such as when $q$ is a Laplace (\cref{sec:laplace}).

\begin{algorithm}[t]
\caption{Variational score matching, used as a benchmark for existing SM objectives.}
\label{alg:var}
\KwIn{model $\ptheta$, variational $\qphi$, $S$ posterior samples, $J$ consecutive updates to $q$}
\While{not converged}{
    \For{k = 1,\dots,J}{
        Update $\vphi \propto \nabla_\vphi\KL[\qphi(\vz|\vx)\|\ptheta(\vz|\vx)]$ w/ repar., or $\nabla_\vphi \FD[\qphi(\vz|\vx)\|\ptheta(\vz|\vx)]$ w/o repar. 
        estimated by $S$ samples from $q$.
    }
    Update $\vtheta\propto\ntheta M_{1,\vphi,\vtheta}(\vx)$ \eqref{eq:M1} estimated by $S$ samples from $q$.
}
\end{algorithm}

Combining the results of FD-based learning and inference 
in \cref{sec:improve_fae,sec:FD_inference}, we summarize the 
benchmark variational SM procedure in \cref{alg:var}.
Since this algorithm is closely related to the ELBO \eqref{eq:vae_ae}, 
we expect their performances to be largely similar without hyperparameter tuning.
Further practical limitations of \cref{alg:var} are discussed in \cref{sec:discuss}, 
and for these reasons, this algorithm is used only to benchmark with other SM objectives 
in VAE models. It is also distinct from FAE \citep{fae}
which failed badly in all our experiments.

\section{Marginal SM objectives do not recover autoencoding losses}\label{sec:related}

There are two closely related variational SM objectives \citep{swersky2011autoencoders, bao2020bi}. 
They optimize approximations to the marginal FD \eqref{eq:marginal_fd} rather than the joint FD, 
with implications that we discuss in detail here.
First, \citet{swersky2011autoencoders} derived an objective that is equivalent to \eqref{eq:marginal_fd} 
up to a $\pi$-dependent constant:
\begin{equation}\label{eq:marginal_fd_exp_ibp}
    \E_{\pi}\left\{\E_\ppost\!\left[ \|\stheta(\vx|\vz)\|_2^2 + \ntheta \cdot \stheta(\vx|\vz)\right]
    - \tfrac{1}{2}\| \E_\ppost[\stheta(\vx|\vz) ] \|_2^2 \right\}.
\end{equation}
For a VAE model, one can replace $\ppost$ with $q$ 
in \eqref{eq:marginal_fd_exp_ibp} to obtain a per-datum loss
\begin{equation}\label{eq:M2}
M_{2,q,\vtheta}(\vx):=
    \Eq\left[ \|\stheta(\vx|\vz)\|_2^2 \!+\! \nx \cdot \stheta(\vx|\vz)\right] - \tfrac{1}{2}\| \Eq\left[\stheta(\vx|\vz) \right] \|_2^2.
\end{equation}
In the special case of Gaussian VAEs, we applied the same technique used 
for \cref{thm:l1} to $M_2$, giving the following autoencoding loss.
\begin{restatable}{proposition}{ltwo}
\label{thm:l2}
The optimal semi-Gaussian VAE \eqref{eq:vae_densities} 
that minimizes $\E_\gD\left[M_{2,\phi, \theta}(\vx) \right]$ 
solves
\begin{gather*}
    \min_{\vtheta,q} \colorhF\!\left(\frac{1}{N}\sum_{n=1}^N\colorLtwo{,q,\vtheta}(\vx_n)\right), \quad
    \colorLtwo{,q,\vtheta}(\vx) = 
        \left\| \vx - \E_{q}\!\left[\vg_\vtheta(\vz))\right] \right\|_2^2 
            + 2\E_{q}\!\left[\left\|\gtheta(\vz)\right\|_2^2\right]
\end{gather*}
where $\colorhF(y) = -d_x^2/(2y)$. 
\end{restatable}
The second term of $\colorLtwo{}$ involves the expected $\ell$-2 norm
of the reconstruction mean, while the second term of $\colorLK{}$ in \eqref{eq:vae_ae} and \eqref{eq:M1_ae} 
is the reconstruction variance.
Thus, the objective proposed by \citet{swersky2011autoencoders} 
applied to semi-Gaussian VAEs overly constrains the reconstruction.

Second, \citet{bao2020bi} inserted the identity
    $\nx\log\ptheta(\vx) = \Esub{\ppost} \left[ \nx\log\ptheta(\vx|\vz)\right]$
into the marginal FD \eqref{eq:marginal_fd}, giving 
\begin{equation}\label{eq:marginal_fd_exp}
    \FD[\pi(\vx)\|\ptheta(\vx)]=\frac{1}{2}\E_\pi\left[ \left\| \nx\log\pi(\vx)-\E_\ppost[\stheta(\vx|\vz)]\right\|_2^2 \right].
\end{equation}
The authors then replaced $\ptheta(\vz|\vx)$ with a variational posterior
$q_{\vphi}(\vz|\vx)$.
We derive its equivalent autoencoding loss below for semi-Gaussian VAEs.
\begin{restatable}{proposition}{mone}
\label{prop:M1}
Approximating $\ptheta(\vz|\vx)$ with $q(\vz|\vx)$,
the marginal FD $\eqref{eq:marginal_fd_exp}$ 
is equal to $\E_\pi[M_{3,q,\vtheta}(\vx)]$,
up to a $\pi$-dependent constant,
where 
\begin{equation}\label{eq:M3}
M_{3,q,\vtheta}(\vx):= 
        \E_q\!\left[
           \vs^q(\vz|\vx)\cdot\stheta(\vx|\vz) + 
           \nx\cdot\stheta(\vx|\vz) 
           \right] + \tfrac{1}{2}\left\|\E_q[\stheta(\vx|\vz)]\right\|_2^2.
\end{equation}
\end{restatable}
In contrast to $M_1$ and $M_2$, the autoencoding 
loss for $M_3$ is not immediately interpretable and is deferred to \cref{sec:ae_objectives}
only for completeness. 

Another common issue of these two objectives above is that replacing the exact posterior 
with a variational $q$ removes its dependence on $\vtheta$. 
\citet{bao2020bi} proposed to update $\vphi$ by $\vtheta$-dependent gradients. 
Nevertheless, this inner optimization is slow in practice and was discarded in the authors' main experiments.
The contribution of this bi-level optimization is thus unclear, which we test empirically in \cref{sec:exp}.
In contrast, the FAE based on the joint FD and \cref{alg:var} based on our benchmark 
objective use a free variational posterior independent of $\vtheta$. 

\begin{figure*}[t!]
    \centering
    \includegraphics[width=\textwidth]{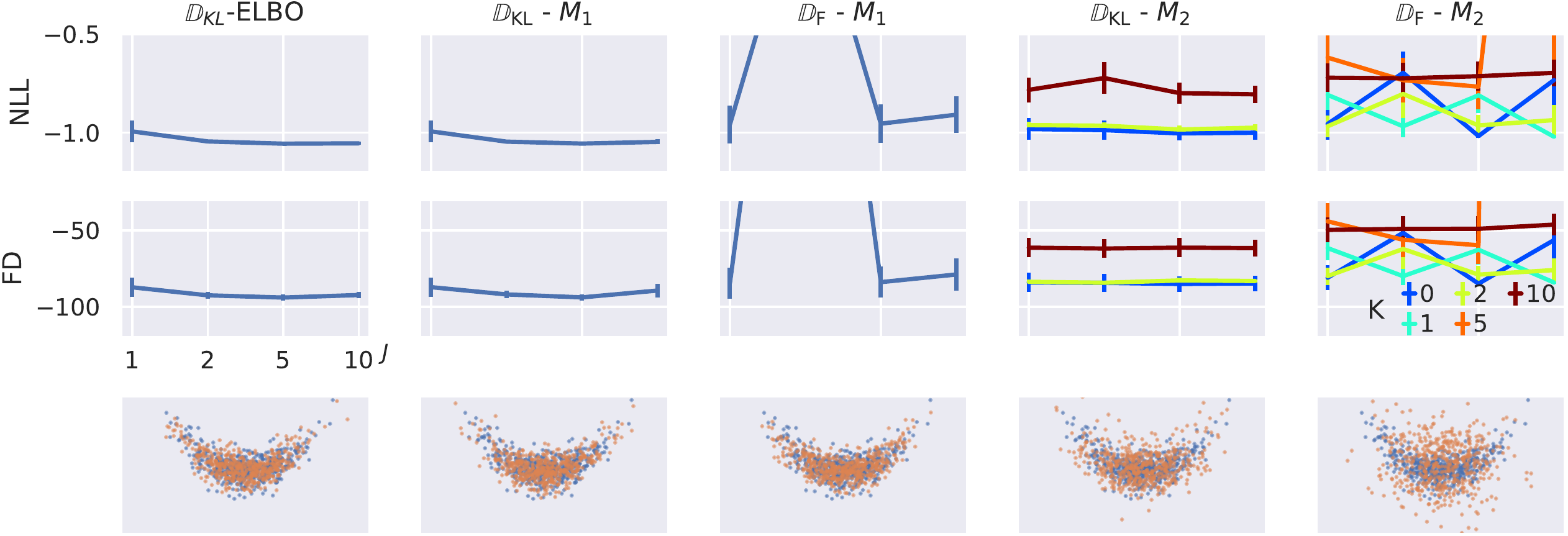}\\
    \vspace{1em}
    \includegraphics[width=\textwidth, trim=0 0 0 0.3cm, clip]{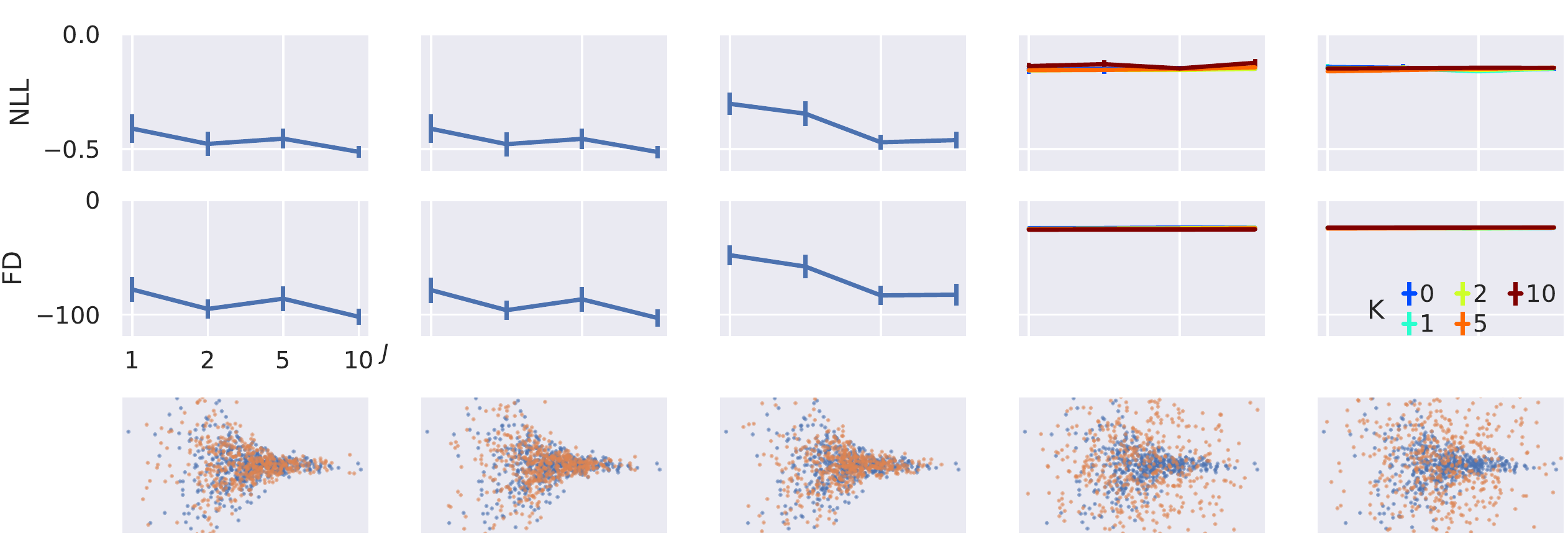}
    \caption{Results of models trained by different objecitves (columns) on two sythetic datasets (top three and bottom three rows). 
    We report the negative log likelihood and FD as we vary the number of independent encoder updates $J$ 
    and the number of bi-level encoder updates $K$, 
    and samples from model with the lowest NLL (orange). Errorbar shows 1 sem from 10 independent initializations.}
    \label{fig:toy}
\end{figure*}

\section{Experiments}\label{sec:exp}

To compare the effects of training objectives on Gaussian VAE models, 
we tested All combinations of the two inference objectives (KLD and FD) and the three SM objectives 
($M$'s), yielding six overall SM objectives. 
The ELBO objectives is used as a baseline.
Note that \citet{bao2020bi} 
used an objective that is equal to $\KL$-$M_3$ in expectation.
Optimizing FD with reparametrized 
gradient, or optimizing the joint FD as in Fisher autoencoder \citep{fae} 
did not give any reasonable results and are excluded for detailed comparison.
The goal of these experiments is to confirm the analyses in previous sections. 
In particular, we expect that $M_1$ is the only learning objective that can produce results similar to ELBO, 
and the other learning objectives are worse.
Thus, we also do not expect the benchmark algorithm or any previous methods applied to VAEs to achieve state-of-the-art performance on any standard metrics.
Details are in \cref{sec:exp_details} and code is available at \url{github.com/kevin-w-li/LatentScoreMatching}.

\subsection{Synthetic datasets}
We first trained Gaussian VAEs composed of fully connected layers (two hidden layers of 30 units) 
on simple 2D datasets. The latent space is $\gR^2$.
For each combined objective, we ran $J\ge1$ gradient updates to the encoder before updating the decoder.
For objectives involving $M_2$ and $M_3$ where bi-level optimization is required, 
we also updated the encoder $K\ge0$ times with gradients that retain its dependence on $\vtheta$. 
The gradient for each update is computed using a minibatch of 1\,000 samples.
Each experimental setting is repeated 10 times. 
After training, the trained models are evaluated by the negative log-likelihood (NLL) 
and score matching loss 
(FD minus a constant) on unseen data, approximated by 
importance sampling using $100\,000$ samples.

The results are shown in \cref{fig:toy}. 
The ELBO objective ($\KL$-ELBO) and $\KL$-$M_1$ produced almost equally good results, 
both the NLL and FD decrease as the number of encoder updates $J$ increases, as expected. 
This confirms our analysis that these two objectives optimize very similar autoencoding losses.
On the other had, $\FD$-$M_1$ is slightly worse,
suggesting that the Fisher divergence is indeed not an ideal objective for learning $\vtheta$.
The objectives involving $M_2$ and $M_3$ could not fit the model well. 
The NLL and FD for $M_3$ objectives were above the limits of the figure axis ranges.
On these two synthetic datasets, we did not observe 
strong positive effects of bi-level optimization. 
But we did observe this effect for much higher $J$, $K$ and network size; see \cref{sec:toy_exp_details} where 
results on other synthetic datasets are also reported.

\begin{figure*}[t!]
    \centering
    \includegraphics[width=\textwidth]{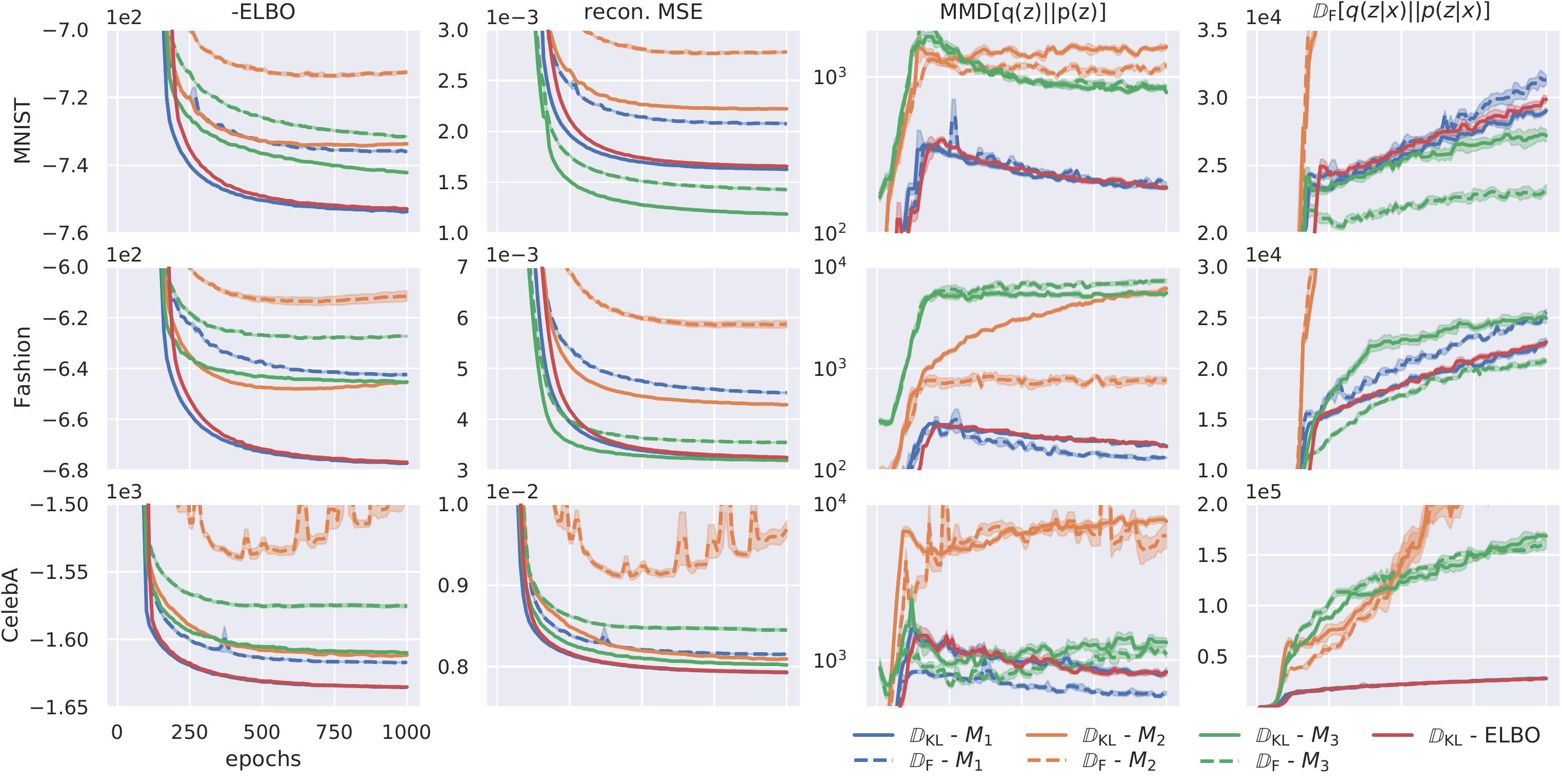}
    \caption{Test metrics (columns) for each dataset (rows). Lower is better. The objectives optimized is 
    shown in the legend. The error bars are estimated based on 5 runs.
    }
    \label{fig:ae}
\end{figure*}
\subsection{Benchmark datasets}\label{sec:benchmarks}

\paragraph{Procedure} We tested the variational SM on the Gaussian VAE model 
MNIST \citep[CC BY-SA 3.0]{LeCunEtAl1998} and FashionMNIST \citep[Fashion][MIT]{fashionmnist}
and CelebA \citep[CC-BY-4.0]{celeb} datasets. All images are resized to $32 \times 32$.
To avoid potentially unbounded gradients as $\gamma\to0$, we added a small Gaussian
noise to the data. This also prevented gross overfitting.
Two neural architectures were tested: ConvNets as defined in DCGAN \citep{radford2015unsupervised} 
and ResNets \citep{he2016deep}. 
To ensure a good posterior approximation, we drew $S=5$ samples from $q$ to estimate expectations
and updated $q(\vz|\vx)$ $J=5$ times for every $\vtheta$ update. 
We did not perform bi-level optimization for large benchmark datasets \citep{bao2020bi}.
Adam with step size $10^{-4}$ was used as the optimization routine, 
and training lasted 1\,000 epochs for each objective on each dataset.
These experiments were run on NVIDIA\textsuperscript\textregistered\ GTX 1080 GPUs.

To make a thorough comparison between the training objectives, we computed the following metrics 
on the test dataset after every 10 epochs: 
a) the \textbf{negative ELBO} as an overall metric;
b) reconstruction mean squared error (\textbf{MSE});
c) the latent maximum mean discrepancy \citep[\textbf{MMD}, ][]{gretton2012kernel} 
    between the aggregate posterior $q(\vz):=\int\pi(\vx)q(\vz|\vx)\ud\vx$ and 
    $p(\vz)$, a global measure of the posterior
    approximation;
and d) the \textbf{average FD} between the approximate and exact posteriors, a local 
    measure of the posterior approximation.
We also computed the \textbf{FID} \citep{fid} and \textbf{KID} \citep{kid}, measuring sample quality, after training.
Note that the latent MMD and the average posterior FD may increase through training 
as the true $\ptheta(\vz|\vx)$ becomes more complicated.

\paragraph{ConvNets}
We focus on the results of ConvNets shown in \cref{fig:ae}.
The benchmark objective $\KL$-$M_1$ gave learning trajectories almost identical   
to the $\KL$-ELBO on all metrics for all datasets.
$\FD$-$M_1$ produced the best negative ELBO compared to the other $\FD$- objectives but 
is still worse than $\KL$-ELBO or $\KL$-$M_1$.
These are again consistent with our analyses.
Comparing between the SM objectives, we found that $M_2$ and $M_3$ underperform $\KL$ -ELBO 
and $M_1$ on negative ELBO and latent MMD. 
$M_2$ gave the highest reconstruction error, which may be the result of \cref{thm:l2}
that predicts over regularization of the reconstruction mean.
$M_3$ produced the lowest reconstruction MSE on the MNIST dataset only, but 
the latent MMD and posterior FD is large on other datasets.

Regarding inference, 
$\FD$-$M_1$ gave the lowest 
latent MMD on Fashion and CelebA, and $\FD$-$M_3$ gave the lowest 
posterior FD on CelebA. However, overall,
methods with FD-based inference fared worse on latent MMD and posterior FD 
than those with $\KL$-based inference. 
Thus, KLD is indeed the preferred objective for variational learning
even when the learning objective is FD, as done in \citet{bao2020bi}. 
The sample qualities are shown in \cref{tab:allexp}. $\KL$+ELBO and the benchmark $M_1$ 
gave some of the best FIDs. 
Again, $\KL$-$M_1$ produced indistinguishable results compared to $\KL$+ELBO, 
but other objectives yielded worse FIDs.

\paragraph{ResNets}
So far, the empirical results are consistent across dataset, networks
and different runs of the same settings. 
What if the VAE uses the more flexible ResNets? 
We focus on the sample quality metrics in (\cref{tab:allexp}) and 
defer more detailed results to \cref{sec:var_exp_details}.
$\KL$-$M_1$ produced better samples than $\KL$-ELBO on all datasets. 
The combination $\FD$-$M_1$ had better sample quality than $\KL$-$M_1$ 
and $\KL$-ELBO (\cref{tab:allexp}) on MNIST. 
Surprisingly, the best sample quality model turned out to be $\FD$-$M_2$, 
although all $\FD$- combinations produced the worst negative ELBO, reconstruction error 
and posterior FD than $\KL$- combinations (\cref{sec:var_exp_details}).
These results suggest strong effects of 
neural architecture on performance when $M_2$ is used. 
In principle, using this objectives for training the decoder would
require a bi-level optimization over $q$, but in practice this 
does not seem necessary when ResNets are used, 
as shown in the experiments here and previously \citep{bao2020bi}.

\paragraph{Sparse representation} 
Finally, to verify whether the objectives can produce sparse latent representation 
, 
we show the histogram of posterior standard deviation (SD) for each 
latent dimension in \cref{sec:var_exp_details}, following \citet{dai2018diagnosing}. 
For models with ResNet architecture, the posterior SDs trained by $M_1$ are concentrated 
around either 0.0 or 1.0 (\cref{fig:ae_sparse_resnet}), consistent with \cref{thm:m1_like_elbo}.  
However, the same pattern is also observed on VAEs trained with the other learning objectives, 
even when the fit of the model is rather poor. 
Further, a sparse representation is also seen in a VAE model with binary latent variables \cref{fig:binary_sparse}.
Therefore, such a representation may be a more general phenomenon caused by factors that is not 
specific to ELBO or $M_1$.
This sparse representation is less prominent using ConvNets \cref{fig:ae_sparse}, especially 
on the more complex CelebA dataset.

\begin{table}[t!]
    \setlength{\tabcolsep}{2.5pt}
    \centering
    \caption{Sample quality (KID $\times~ 10^3$) of model trained by different methods. $\sK:=\KL$ and $\sF:=\FD$.}\label{tab:allexp}
    {
    \begin{tabular}{ll*{7}{c}|ccc}
        \toprule
        \multicolumn{2}{l}{\textbf{Models}} & \multicolumn{7}{c|}{\textbf{ConvNets} (5 runs, s.e $\le5\%$)} & \multicolumn{3}{c}{\textbf{ResNets}}\\
        \multicolumn{2}{l}{\textbf{Objectives}} & {\color{tab:red}$\sK$-ELBO} & {\color{tab:blue}$\sK$-$M_1$} & {\color{tab:blue}$\sF$-$M_1$} & {\color{tab:orange}$\sK$-$M_2$} & {\color{tab:orange}$\sF$-$M_2$} & {\color{tab:green}$\sK$-$M_3$} & {\color{tab:green}$\sF$-$M_3$} & {\color{tab:red}$\sK$-ELBO}  & {\color{tab:blue}$\sK$-$M_1$} & {\color{tab:blue}$\sF$-$M_1$} \\
        \hline
        \multirow{2}{*}{MNIST}  & FID  & 5.27  & 5.22 & 5.87 & 21.1 & 21.1 & 12.2 & 13.5 & 5.58 & 5.24  &  2.93 \\
                                & KID  & 63.7  & 60.8 & 68.8 & 402  & 417  & 161  & 195  & 96.8 & 90.8  &  54.7 \\
        \hline
        \multirow{2}{*}{Fashion}& FID  & 5.60  & 5.47 & 5.26 & 29.1 & 25.2 & 19.2 & 24.0 & 4.89 & 4.63  &  4.86 \\
                                & KID  & 58.2  & 55.4 & 55.1 & 465  & 360  & 300  & 403  & 81.1 & 73.5  &  79.4 \\
        \hline
        \multirow{2}{*}{Celeb}  & FID  & 144   & 145  & 153  & 170  & 172  & 157  & 178  & 55.4 & 54.6  &  59.2 \\
                                & KID  & 160   & 161  & 173  & 156  & 166  & 165  & 196  & 56.9 & 56.1  &  58.6 \\
        \bottomrule
    \end{tabular}
    }
\end{table}

\section{Discussion}\label{sec:discuss}

We performed an analytical and empirical comparison
between existing score matching objectives \citep{swersky2011autoencoders,bao2020bi,fae}. 
Despite their excellent empirical results on image generation, the disparate  
model classes and training algorithms employed obscure the contributions from the training objectives. 
Here, we ask how these objectives perform in Gaussian VAEs that can be trained by 
maximum-likelihood and score matching. 
We found that minimizing the joint Fisher divergence \citep{fae} 
resulted in a significant bias in the learned distribution, but a simple modification 
gives rise to an objective $M_1$ that resembles the ELBO on Gaussian VAEs. 
When combined with KLD-based inference, 
this objective yielded very similar performance with ELBO on different model 
architectures and datasets.
Other learning objectives derived from the marginal Fisher divergence \citep{swersky2011autoencoders,bao2020bi} 
correspond to less effective autoencoding losses and require an expensive bi-level optimization in principle.
They failed to learn simple synthetic datasets but appear more competent with 
more flexible neural architectures on real datasets. 
In addition, like the ELBO, the posterior FD can act as a regularizer 
on the encoded mean, but optimization on this objective can lead to poor local optima. In practice, 
KLD-based inference yielded better results over FD-based inference.

It is possible that our comparison ignored other important properties of these objectives. 
For example, the objectives that failed to perform well may have enormous estimation variances 
than the benchmark $M_1$, which resulted in less efficient learning in practice. 
Further, the benchmark algorithm has limitations that are worth mentioning. 
First, the learning objective $M_1$ is not applicable to general jointly energy-based models.
Second, the similarity between $M_1$ and the ELBO may not be so obvious for 
non-Gaussian likelihoods; other SM objectives 
for certain likelihoods \citep{raphan2007learning} may connect $M_1$ and ELBO in more general cases.
Despite these, our main message is clear: on the Gaussian VAE model, 
previous variational score matching algorithms often struggled to deliver
satisfactory performance compared to variational maximum-likelihood or the benchmark score 
matching algorithm, which have more predictable behaviors.
While this work does not in any way undermine the theoretical contributions made in those studies, 
it calls into question the contributions of their proposed \emph{training objectives} to the empirical performances,
and we cannot ignore the effects of detailed factors in the model and experiment.

\printbibliography

\medskip

\clearpage
\onecolumn
\appendix

\begin{center}
\Large {\textbf{\mytitle:\\Supplementary material}}
\end{center}

\section{Proofs and derivations}
\subsection{FD between joint distributions}\label{sec:FD_two_terms_deriv}

Starting from the joint Fisher divergence, we have
\begin{align*}
     \FD[&q(\vz|\vx)\pi(\vx)\|\ptheta(\vz,\vx)]:= 
     \frac{1}{2}\Esub{q\pi}\left[ 
            \left\| \nabla_{\vz,\vx} \log \frac{q(\vz|\vx)\pi(\vx)}{\ptheta(\vz)\ptheta(\vx|\vz)} \right\|_2^2 
            \right]\\
       &=   \frac{1}{2}\Esub{q \pi}\left[
            \left\| \nabla_{\vz} \log \frac{q(\vz|\vx)\pi(\vx)}{\ptheta(\vz)\ptheta(\vx|\vz)} \right\|_2^2 +
            \left\| \nabla_{\vx} \log \frac{q(\vz|\vx)\pi(\vx)}{\ptheta(\vz)\ptheta(\vx|\vz)} \right\|_2^2\right]  \\
       &=   \frac{1}{2}\Esub{q \pi}\left[
            \left\| \nabla_{\vz} \log \frac{q(\vz|\vx)\pi(\vx)}{\ptheta(\vz|\vx)\ptheta(\vx)} \right\|_2^2 +
            \left\| \nabla_{\vx} \log \frac{q(\vz|\vx)\pi(\vx)}{\ptheta(\vz|\vx)\ptheta(\vx)} \right\|_2^2 \right] \\
       &=   \underbrace{\frac{1}{2}\Esub{q \pi}\left[
            \left\| \nabla_{\vz} \log \frac{q(\vz|\vx)}{\ptheta(\vz|\vx)} \right\|_2^2
            \right]}_{\tfrac{1}{2}\Esub{\pi}\FD[q(\vz|\vx)\|\ptheta(\vz|\vx)]} +
            \Esub{q\pi}\left[
            \left\| \nabla_{\vx} \log \frac{\pi(\vx)}{\ptheta(\vx)} + 
                \nabla_\vx \log \frac{q(\vz|\vx)}{\ptheta(\vz|\vx)}\right\|_2^2 \right],\\
\end{align*}
which is \eqref{eq:FD_bounds}. The second line is obtained by 
separating the squared $\ell$-2 norm of derivatives over $\vz$ and $\vx$ into two 
separate norms. The third line holds by applying the Bayes rule to the denominators
in the logarithms.
In the last line, we used the fact that $\nz\log\ptheta(\vx)=0$ and 
$\nz\log\pi(\vx)=0$.

Thus, the joint FD decomposes into a sum of two terms that are, respectively, related to the partial 
derivatives w.r.t\  $\vz$ and $\vx$. Below, we simplify the second term to get rid of the dependence on 
the unknown $\nx\log\pi(\vx)$. One can show that the second term is equal to 
\[
\Jqtp :=\Esubscore{q(\vz|\vx)}{\nx\log\ptheta(\vx|\vz) - \nx\log q(\vz|\vx) - \nx\log\pi(\vx)}.
\]
Then one has that 
\begin{align*}
\Esub{\pi}\left[\Jqtp\right]&:=
    \frac{1}{2}\Esubscore{q\pi}{\nx\log\ptheta(\vx|\vz) - \nx\log q(\vz|\vx) - \nx\log\pi(\vx)}\\
    &=\frac{1}{2}\Esub{q\pi}\left[\|\nx\log\ptheta(\vx|\vz)\|_2^2 + \| \nx\log q(\vz|\vx) \|_2^2 + 
                       \|\nx\log \pi(\vx)\|_2^2 \right]\\
    &\qquad + \Esub{q\pi}\big[ 
                                {\color{red}\nx\log q(\vz|\vx) \cdot \nx\log \pi(\vx) }
                                - {\color{green}\nx\log\ptheta(\vx|\vz) \cdot \nx\log \pi(\vx)}.\\
                            & \qquad - {\color{blue} \nx\log q(\vz|\vx) \cdot \nx\log\ptheta(\vx|\vz) }
                                \big]\numberthis\label{eq:cross-terms}
\end{align*}
We will now simplify the first and second cross-terms (colored). The first one is zero. (The derivation by \citet{fae} suggests that this term may be nonzero but depends only on $\pi$, 
which is an inconsequential error.)
\begin{align*}
\Esub{q\pi}\left[ {\color{red}\nx\log q(\vz|\vx) \cdot \nx\log \pi(\vx)}  \right] 
        = \Esub{\pi}\left[\Esub{q(\vz|\vx)}\left[ \nx\log q(\vz|\vx)\right]\cdot \nx\log \pi(\vx) \right] = 0.
\end{align*}
The second cross-term of \eqref{eq:cross-terms} is simplified as
\begin{align*}
&\qquad \Esub{q\pi}\left[ {\color{green}\nx\log \ptheta(\vx|\vz) \cdot \nx\log \pi(\vx)}  \right] \\
&= \Esub{\pi}\left[ \Esub{q(\vz|\vx)}\left[\nx\log\ptheta(\vx|\vz)\right] \cdot \nx\log\pi(\vx) \right]\\
&\stackrel{(1)}{=}-\Esub{\pi}\left[\nx\cdot \Esub{q(\vz|\vx)}\left[\nx\log\ptheta(\vx|\vz)\right] \right]\\
&\stackrel{(2)}{=} - \Esub{q\pi} \left[ {\color{blue}\nx \log q(\vz|\vx)\cdot \nx\log \ptheta(\vx|\vz)}\right]
   - \Esub{q\pi} \left[\nx \cdot \nx\log \ptheta(\vx|\vz)  \right],\\
\end{align*}
where (1) follows from \eqref{eq:ibp} and (2) uses the score trick.
Substituting back to \eqref{eq:cross-terms} cancels the blue term, and we arrive at $\Esubscore{\pi}{\Jqtp}$
\begin{align*}
    &=\Esub{q\pi}\Bigg[\frac{1}{2}\| \nx\log\ptheta(\vx|\vz)\|_2^2 + \nx \cdot \nx\log \ptheta(\vx|\vz)
        +\frac{1}{2} \| \nx\log q(\vz|\vx) \|_2^2 \\
    & \qquad\qquad +\frac{1}{2}\|\nx\log \pi(\vx)\|_2^2 \Bigg]\\
    &=\Esub{\pi}\left[\underbrace{\Esub{q}\left[\frac{1}{2}\| \nx\log\ptheta(\vx|\vz)\|_2^2 
                        + \nx \cdot \nx\log \ptheta(\vx|\vz)  
        +\frac{1}{2} \| \nx\log q(\vz|\vx) \|_2^2\right]}_{M_{1,q,\vtheta}(\vx)} + C_\pi\right].\\
\end{align*}
where the $\pi$-dependent constant 
\begin{align}\label{eq:pi-const}
    C_\pi:= \frac{1}{2} \left\|\E_{\pi(\vx)}[\nx\log\pi(\vx)]\right\|_2^2,
\end{align}

\subsection{Autoencoding objectives for semi-Gaussian VAEs}\label{sec:ae_objectives}
Here, we show that the objectives $M_1$ to $M_3$ reduce to autoencoding objectives 
similar to \eqref{eq:vae_ae} for semi-Gaussian VAEs where
$p(\vx|\vz)=\gN(\gtheta(\vz), \gamma \mI)$.
We will use the following identities for semi-Gaussian VAEs extensively in our derivations
\begin{gather}
    \stheta(\vx|\vz) = -\frac{1}{\gamma}(\vx-\vg_\vtheta(\vz))\label{eq:gaussian_score}\\
    \left\|\Esub{q}\left[\stheta(\vx|\vz)\right]\right\|_2^2 = 
        \frac{1}{\gamma^2}\left\|\vx-\Eq\left[\vg_\vtheta(\vz)\right]\right\|_2^2\label{eq:gaussian_norm_E}\\
    \Eq\left[\left\|\stheta(\vx|\vz)\right\|_2^2\right]
        = \frac{1}{\gamma^2}\Eq\left[\left\|\vx-\vg_\vtheta(\vz)\right\|_2^2\right]
        = \frac{1}{\gamma^2}\left[\left\|\vx-\Eq\left[\vg_\vtheta(\vz)\right]\right\|_2^2 
                                + \Tr\sC_q[\gtheta(\vz)]\right]
            \label{eq:gaussian_E_norm}\\
    \nx\cdot\stheta(\vx|\vz)= -\frac{d_x}{\gamma}\label{eq:gaussian_second}
\end{gather}

In addition, one can check the following identity holds for any $a>0$ and $b\in\sR$
\begin{equation}\label{eq:hyper}
    \min_\gamma \left\{\frac{a}{2\gamma^2} - \frac{b}{\gamma}\right\} = -\frac{b^2}{2a}
\end{equation}

\lone*
\begin{proof}
Since $q$ is fixed, $M_1$ is equal to the following up to a $q$-dependent constant
\begin{align*}
    \tilde{M}_{1,q,\vtheta}(\vx)
       & := \Eq\left[\frac{1}{2}\left\| \stheta(\vx|\vz) \right\|_2^2 + \nabla_\vx\cdot\stheta(\vx|\vz)\right]  \\
       & = \Eq\left[
            \frac{1}{2\gamma^2}\left(\left\|\vx-\Eq\left[\vg_\vtheta(\vz)\right]\right\|_2^2 
            + \sum_{j=1}^{d_z} \sV_q[\gtheta(\vz)_j]  \right)\right] - \frac{d_x}{\gamma}\\
       & = \frac{1}{2\gamma^2}\left\{
            \left\|\vx-\Eq\left[\vg_\vtheta(\vz)\right]\right\|_2^2 + \sum_{j=1}^{d_z} \sV_q[\gtheta(\vz)_j] \right\} 
            - \frac{d_x}{\gamma}\\
    \E_\gD\left[\tilde{M}_{1,q,\vtheta}(\vx)\right] &= 
       \frac{1}{2\gamma^2}\left\{
            \frac{1}{N}\sum_{n=1}^N
                    \left\|\vx_n-\E_{q(\vz|\vx_n)}\left[\vg_\vtheta(\vz)\right]\right\|_2^2 
                        + \sum_{j=1}^{d_z} \sV_{q(\vz|\vx_n)}[\gtheta(\vz)_j] 
            \right\} - \frac{d_x}{\gamma}\\
\end{align*}
The optimal VAE has an optimal $\vgamma$. Using \eqref{eq:hyper} to optimize out $\gamma$, we conclude that the optimal $\vtheta$ is the solution to
\begin{equation}
    h\left(
            \frac{1}{N}\sum_{n=1}^N
                    \left\|\vx_n-\E_{q(\vz|\vx_n)}\left[\vg_\vtheta(\vz)\right]\right\|_2^2 
                        + \sum_{j=1}^{d_z} \sV_{q(\vz|\vx_n)}[\gtheta(\vz)_j] 
        \right)
\end{equation}
where $h(y)=-d_x^2/(2y)$
\end{proof}

\ltwo*
\begin{proof}
We expand $M_2$ using \eqref{eq:gaussian_score}-\eqref{eq:gaussian_second} as
\begin{align*}
M_{2,q,\vtheta}(\vx)&:=
    \Eq\left[ \|\stheta(\vx|\vz)\|_2^2 \!+\! \nx \cdot \stheta(\vx|\vz)\right]\!-\! \frac{1}{2}\| \Eq\left[\stheta (\vx|\vz)\right] \|_2^2.\\
    &=\frac{1}{\gamma^2}\Eq\left[\left\|\vx-\vg_\vtheta(\vz)\right\|_2^2\right] 
        -  \frac{1}{2\gamma^2}\|\vx-\Eq\left[\vg_\vtheta(\vz)\right]\|_2^2 - \frac{d_x}{\gamma}\\
    &=\frac{1}{2\gamma^2}\left\{\left\| \vx - \Eq\left[ \gtheta(\vz)\right] \right\|_2^2     
        + 2\Eq\left[\left\|\gtheta(\vz)\right\|_2^2\right] \right\}
        - \frac{d_x}{\gamma}\\
\E_\gD \left[ M_{2,q,\vtheta}(\vx) \right] &= \frac{1}{N}\sum_{n=1}^N M_{2,q,\vtheta}(\vx_n)\\
    &=\frac{1}{2\gamma^2}\left\{\frac{1}{N}\sum_{n=1}^N
                    \left\| \vx_n - \E_{q(\vz|\vx_n)}\left[ \gtheta(\vz)\right] \right\|_2^2     
                    + 2\E_{q(\vz|\vx_n)}[\left\| \gtheta(\vz) \right\|_2^2] \right\}
        - \frac{d_x}{\gamma}\\
\end{align*}
We then take the minimum w.r.t.\ $\gamma$ and using \eqref{eq:hyper} to obtain the loss
\begin{equation}
    h\left(
    \frac{1}{N}\sum_{n=1}^N
                    \left\| \vx_n - \E_{q(\vz|\vx_n)}\left[ \gtheta(\vz)\right] \right\|_2^2     
                    + 2\E_{q(\vz|\vx_n)}[\left\| \gtheta(\vz) \right\|_2^2] 
        \right)
\end{equation}
where $h(x)=-d_x^2/(2x)$
\end{proof}

\begin{restatable}{proposition}{lthree}
\label{thm:l3}
The optimal semi-Gaussian VAE \eqref{eq:vae_densities} 
that minimizes $\E_\gD\left[M_{3,\phi, \theta}(\vx) \right]$ 
is the solution to
\begin{gather*}
    \min_{\vtheta,q} h\left(\frac{1}{N}\sum_{n=1}^N\gL_{3,q,\vtheta}(\vx_n)\right)\\
    \gL_{3,q,\vtheta}(\vx) = 
        g(\left\| \vx_n - \E_{q(\vz|\vx_n)}\left[\gtheta(\vz)\right] \right\|_2^2\\
g(x) = -\frac{1}{2x}\left(\frac{1}{N}\sum_{n=1}^N \Eq\left[\vs^q(\vz|\vx_n)\cdot(x_n-\gtheta(\vz)) \right]+ d_x\right)^2
\end{gather*}
\end{restatable}
\begin{proof}

We expand $M_3$ using \eqref{eq:gaussian_norm_E} and \eqref{eq:gaussian_second} as
\begin{align*}
M_{3,q,\vtheta}(\vx)&:= 
    \frac{1}{2}\left\|\Eq\left[\stheta(\vx|\vz)\right]\right\|_2^2  
        + \Eq\left[ \vs^q(\vz|\vx)\cdot\stheta(\vx|\vz) + 
           \nx\cdot\stheta (\vx|\vz)
            \right].\\
    &= \frac{1}{2\gamma^2}\left\|\vx-\Eq\left[\gtheta(\vz) \right] \right\|_2^2
            - \frac{1}{\gamma}\left(\Eq\left[\vs^q(\vz|\vx)\cdot(\vx-\gtheta(\vz)) \right] 
                                    + d_x\right)\\
\E_\gD\left[M_{3,q,\vtheta}(\vx)\right]
    &= \frac{1}{2\gamma^2}\left\{\frac{1}{N}\sum_{n=1}^N \left\|\vx_n-\E_{q}\left[\gtheta(\vz) \right] \right\|_2^2\right\}\\
    & \qquad - \frac{1}{\gamma}\left\{\frac{1}{N}
                \sum_{n=1}^N \E_{q}\left[\vs^q(\vz|\vx_n)\cdot(\vx_n-\gtheta(\vz)) \right]
                                    + d_x\right\}
\end{align*}
We conclude the statement using \eqref{eq:hyper}.

\end{proof}

\subsection{Score matching objectives}\label{sec:sm_objectives}
We will use the following due to integration-by-parts when $\pi(\vx)\to 0$ for sufficiently large $\vx$.

\begin{align*}
    &\Esub{\pi(\vx)}\left[ \nx\log \pi(\vx) \cdot \vf(\vx)\right]
    = \int \nx \pi(\vx) \cdot \vf(\vx)\ud \vx
    = [\pi(\vx) \cdot \vf(\vx)]_{\vx\to-\infty}^{\vx\to\infty} - \E_\pi(\vx)\left[ \nx\cdot\vf(\vx)\right]\\
    =& - \Esub{\pi(\vx)}\left[ \nx\cdot\vf(\vx)\right]\numberthis\label{eq:ibp}
\end{align*}
\mone*
\begin{proof}
We expand $\eqref{eq:marginal_fd_exp}$ while approximating the  posterior with $q$.
\begin{align*}
    &\frac{1}{2}\Esub{\pi(\vx)}\left[\| \nx\log\pi(\vx)  - \Eq\left[\stheta(\vx|\vz)\right]\|_2^2 \right]\\
    =&\frac{1}{2}\Esub{\pi(\vx)}\left[ \left\| \Eq\left[\stheta(\vx|\vz) \right] \right\|_2^2 \right]
        - \Esub{\pi(\vx)}\left[\nx\log\pi(\vx) \cdot \Esub{q(\vz|\vx)}\left[\stheta(\vx|\vz)\right] \right]
        + \frac{1}{2} \left\|\Esub{\pi(\vx)}\left[\nx\log\pi(\vx)\right]\right\|_2^2\\
    =&\frac{1}{2}\Esub{\pi(\vx)}\left[ \left\| \Eq\left[\stheta(\vx|\vz) \right] \right\|_2^2 \right]
        + \Esub{\pi(\vx)}\left[ \nx\cdot \Esub{q(\vz|\vx)}\left[\stheta(\vx|\vz)\right] \right]
        + C_\pi\\
    =&\frac{1}{2}\Esub{\pi(\vx)}\left[ \left\| \Eq\left[\stheta(\vx|\vz) \right] \right\|_2^2 \right]
        + \Esub{\pi(\vx)}\left[\int\nx q(\vz|\vx)\cdot \stheta(\vx|\vz)\ud\vx
            + \Eq\left[\nx\cdot\stheta(\vx|\vz)\right]\right] 
        + C_\pi\\
    \stackrel{(a)}{=}&\Esub{\pi(\vx)}\left[ \frac{1}{2}\left\| \Eq\left[\stheta(\vx|\vz) \right] \right\|_2^2 
        + \Esub{q(\vz|\vx)}\left[\nx\log q(\vz|\vx) \cdot \stheta(\vx|\vz)
            + \nx\cdot\stheta(\vx|\vz) \right] \right]
        + C_\pi\\
    =&\E_{\pi(\vx)}\left[ M_{3,q,\vtheta}(\vx) \right] + C_\pi
\end{align*}
where we have used the score approximator for the integral in equality (a).
\end{proof}

\subsection{Non-vanishing posterior fisher divergence in Gaussian VAEs}\label{sec:nonzerofd_proof}
The Fisher divergence between the approximate and exact posterior is
\begin{align*}
&\quad\FD[q(\vz|\vx)\|\ptheta(\vz|\vx)])\\
    &= \int q(\vz|\vx) \left\| 
        \nabla_{\vz} \log q(\vz|\vx) - \nabla_\vz \log \ptheta(\vz) - \nabla_\vz\log \ptheta(\vx|\vz) \right\|_2 \ud\vx\ud\vz
\end{align*}
For Gaussian VAEs, we substitute in the Gaussian densities in \eqref{eq:vae_densities}.
To avoid unnecessary notational clutter, we will suppress the arguments to functions subscripts 
when no confusion arises.
\begin{align*}
\FD[q(\vz|\vx)\|\ptheta(\vz|\vx)]
    &= \int q(\vz|\vx) \left\|
    \mLambda(\vx)(\vz-\vmu(\vx)) - \vz  + 
    \nabla_\vz\vg_\vtheta(\vz)\frac{\vx-\vg_\vtheta(\vz)}{\gamma}
    \right\|_2^2 \ud\vx\ud\vz\\
    &= \int q(\vz|\vx) \left\|
    \vd_{\vtheta}(\vx,\vz) 
    \right\|_2^2 \ud\vx\ud\vz\numberthis\label{eq:FD_post_bound}
\end{align*}
where we have defined 
\begin{equation}\label{eq:distance_vector}
    \vd_{\vtheta}(\vx,\vz) 
    := \mLambda(\vx)(\vz-\vmu(\vx)) - \vz
        + \nabla_\vz\vg_\vtheta(\vz)\frac{\vx-\vg_\vtheta(\vz)}{\gamma}
\end{equation}
Assume that the posterior will be relatively tight around the posterior mean $\vmu(\vx)$,
we expand the last term in the $l$-2 norm by Taylor expansion
\begin{align*}
\nabla_\vz\vg_\vtheta(\vz)(\vx-\vg_\vtheta(\vz))
&= 
    \left.\left[
        \nabla_\vz\vg_\vtheta(\vz)(\vx-\vg_\vtheta(\vz))
    \right]\right|_{\vmu} +\nabla_\vz\!\left.\left[\nabla_\vz\vg_\vtheta(\vz)(\vx-\vg_\vtheta(\vz))\right]\right|_{\vmu}(\vz-\vmu)+ o(\vz^2)\\
&=
    \nabla_\vz\vg_\vtheta(\vz)|_{\vmu}(\vx-\vg_\vtheta(\vmu)) \\
&\qquad +\left[
        \nz\nabla_\vz\vg_\vtheta(\vz)(\vx-\vg_\vtheta(\vmu)) 
        - (\nabla_\vz\vg_\vtheta(\vz)\nabla_\vz\vg_\vtheta(\vz)^\intercal)\rvert_{\vmu}
    \right](\vz-\vmu) \\
&\qquad + o_2((\vz-\vmu)),
\end{align*}
where $o_2(\vx)$ denotes terms that are 2nd or higher order in $\vx$. 
When the model has been trained to describe the data manifold by the latent space, $\vmu$ and 
$\vg_\vtheta$ are inverse of one another, then $\vx-\vg_\vtheta(\vmu(\vx))=0$, we have

$$
    \nabla_\vz\vg_\vtheta(\vz)\frac{\vx-\vg_\vtheta(\vz)}{\gamma} = 
        - \underbrace{\frac{1}{\gamma}(\nabla_\vz\vg_\vtheta(\vz)\nabla_\vz\vg_\vtheta(\vz)^\intercal)\rvert_{\vz=\vmu(\vx)}}_{\mG_{\vtheta}(\vx)}
            (\vz-\vmu(\vx)) + o_2(\vz-\vmu).
$$
Inserting this into the $l$-2 norm of \eqref{eq:distance_vector} gives
\begin{align*}
    \vd_{\vtheta}(\vx,\vz) 
    &= \mLambda(\vx)(\vz-\vmu(\vx)) - \vz 
        - \mG_{\vtheta}(\vx)(\vz-\vmu(\vx)) + o(\vz^2) \\
    &=  \left[\mLambda(\vx) - \mI - \mG_{\vtheta}(\vx)\right]\vz -
        \left[\mLambda(\vx) - \mG_{\vtheta}(\vx)\right] \vmu(\vx) + o_2(\vz-\vmu).
\end{align*}
Since $\vz\sim q(\vz|\vx)$, we can rewrite $\vz=\vmu(\vx)+\mLambda(\vx)^{-\frac{1}{2}}\vepsilon$, 
where $\vepsilon\sim\gN(0,\mI)$, which gives
\begin{align*}
    \vd_{\vtheta}(\vx,\vz) 
    = 
        - \vmu(\vx) 
        + \left[\mLambda(\vx) - \mI - \mG_{\vtheta}(\vx)\right]\mLambda(\vx)^{-\frac{1}{2}}\vepsilon
        + o_2(\mLambda(\vx)^{-\frac{1}{2}}\vepsilon).
\end{align*}
Inserting this back to \eqref{eq:FD_post_bound} gives
\begin{align*}
\FD[q(\vz|\vx)\|\ptheta(\vz|\vx)]
&\approx
    \|\vmu(\vx)\|_2^2 \\
&\qquad + \Tr\left[\left(\mLambda(\vx) - \mI - \mG_{\vtheta}(\vx)\right)
        \mLambda(\vx)^{-1} 
        \left(\mLambda(\vx) - \mI - \mG_{\vtheta}(\vx)\right)    \right].
\end{align*}
Using standard multivariate calculus, it is straightforward to show that 
the last term reaches the minimum of 0 when $\mLambda = \mI + \mG_\vtheta(\vx)$.
(The other solution $\mLambda= - (\mI + \mG_\vtheta(\vx))$ is not positive-semidefinite.)

\subsection{Issues of Fisher divergence in inference}

\subsubsection{Unparametrized FD gradients can resemble KLD gradients}\label{sec:fd_grad_like_kl}

Here, we compare the gradients of KLD and FD between two univariate Gaussian 
distributions. For FD, we will also compute the expression for the biased gradient
obtained using unparametrized samples. 

Consider two Gaussians defined as 
\begin{gather}
    p_1(x) = \gN(x;m_1,s_1^2), \qquad p_2(x)=\gN(x;m_2, s_2^2),
\end{gather}
And our goal is to optimize $p_1$ to be close to $p_2$. The KLD between them is 
\begin{align*}
    \KL[p_1\|p_2] &= \int q(x)\log \frac{q(x)}{p(x)}\ud x \\
           &= -\frac{1}{2} \int p_1(x) \left[ \frac{(x-m_1)^2}{s_1^2} + \log(2\pi s_1^2) -
                                        \frac{(x-m_2)^2}{s_2^2} - \log(2\pi s_2^2)\right]\ud x   \\
           &= -\frac{1}{2} \left[\frac{s_1^2}{s_1^2} + \log(2\pi s_1^2) -
                                        \frac{m_1^2+s_1^2-2m_1m_2+m_2^2}{s_2^2} - \log(2\pi s_2^2)\right] \\  
           &= -\frac{1}{2} \left[ 2\log\left(\frac{s_1}{s_2}\right) -
                                        \frac{(s_1^2-s_2^2) + (m_1-m_2)^2}{s_2^2} \right] \\  
\end{align*}
To optimize the KL w.r.t\ $m_1$ and $s_1$, we can compute the gradients as
\begin{align}\label{eq:gaussian_kl_grad}
    \frac{\partial}{\partial m_1}\KL[p_1\|p_2] 
            &= \frac{m_1-m_2}{s_2^2}, &
    \frac{\partial}{\partial s_1}\KL[p_1\|p_2] 
            &= \frac{s_1^2-s_2^2}{s_1s_2^2}.
\end{align}

We now derive the expression for the FD gradient. Starting from the FD itself:
\begin{align*}
    \FD[p_1\|p_2]   &= \int q(x)\left| \frac{\ud}{\ud x}\log q(x) - \frac{\ud}{\ud x}\log p(x)\right|^2\ud x \numberthis\label{eq:fisher_integral}\\
                    &= \int q(x)\left| \frac{x-m_1}{s_1^2} - \frac{x-m_2}{s_2^2} \right|^2\ud x \\
                    &= \int q(x)\left[\frac{(x-m_1)^2}{s_1^4} - \frac{2(x-m_1)(x-m_2)}{s_1^2s_2^2} + \frac{(x-m_2)^2}{s_2^4} \right]\ud x \\
                    &= \frac{s_1^2}{s_1^4} - 2\frac{m_1^2+s_1^2-m_1^2-m_1m_2+m_1m_2}{s_1^2s_2^2} + \frac{m_1^2+s_1^2-2m_1m_2+m_2^2}{s_2^4} \\
                    &= \frac{1}{s_1^2} - \frac{2}{s_2^2} + \frac{s_1^2+(m_1-m_2)^2}{s_2^4}. \\
\end{align*}
Then the derivatives w.r.t\ $m_1$ and $s_1$ are 
\begin{align}\label{eq:gaussian_fisher_grad}
    \frac{\partial}{\ud m_1}\FD[p_1\|p_2]   &= \frac{2(m_1-m_2)}{s_2^4}, &
    \frac{\partial}{\ud s_1}\FD[p_1\|p_2]   &= \frac{2(s_1^4-s_2^4)}{s_1^3s_2^4}.
\end{align}

Finally, we derive the biased gradient that ignores the dependency on the parameters 
through the $q$ over which the expectation \eqref{eq:fisher_integral} is computed. 
This can be done by evaluating the parameter derivatives first before taking the expectation.
\begin{align*}
    \nabla_v^{b}\FD[p_1\|p_2]   &:= \int q(x) \nabla_v d(x)\ud x, \quad  v\in\{m_1, s_1\} \\
                        d(x)&:=\left| \frac{\ud}{\ud x}\log q(x) - \frac{\ud}{\ud x}\log p(x)\right|^2 = \frac{x-m_1}{s_1^2} - \frac{x-m_2}{s_2^2} 
\end{align*}
\begin{align*}
    \frac{\partial}{\partial m_1}d(x) &= - \frac{2}{s_1^2} \left[ \frac{x-m_1}{s_1^2} - \frac{x-m_2}{s_2^2} \right] &
    \frac{\partial}{\partial s_1}d(x) &= - \frac{4(x-m_1)}{s_1^3} \left[ \frac{x-m_1}{s_1^2} - \frac{x-m_2}{s_2^2} \right]
\end{align*}
Now we evaluate the expectations to get 
\begin{align}\label{eq:gaussian_biased_fisher_grad}
    \frac{\partial^b}{\partial{m_1}}\FD[p_1\|p_2] &= \frac{2(m_1-m_2)}{s_1^2s_2^2} &
    \frac{\partial^b}{\partial{s_1}}\FD[p_1\|p_2] &= \frac{4(s_1^2-s_2^2)}{s_1^3s_2^2}
\end{align}
Thus, combining \eqref{eq:gaussian_kl_grad}, \eqref{eq:gaussian_fisher_grad} and \eqref{eq:gaussian_biased_fisher_grad}, 
we have
\begin{align*}
    \frac{\partial}{\partial{m_1}}\FD[p_1\|p_2] &= \frac{2}{s_2^2}\frac{\partial}{\partial{m_1}}\KL[p_1\|p_2], &
    \frac{\partial}{\partial{s_1}}\FD[p_1\|p_2] &= \frac{2}{s_1^2s_2^2}\frac{\partial}{\partial{s_1}}\KL[p_1\|p_2]; \\
    \frac{\partial^b}{\partial{m_1}}\FD[p_1\|p_2] &= \frac{2}{s_1^2}\frac{\partial}{\partial{m_1}}\KL[p_1\|p_2], &
    \frac{\partial^b}{\partial{s_1}}\FD[p_1\|p_2] &= \frac{4}{s_1^2}\frac{\partial}{\partial{s_1}}\KL[p_1\|p_2].
\end{align*}
The biased FD gradient is equal to the KLD gradient up to a factor $\propto s_1^{-2}$.  
The unbiased gradient is also a scaled version of the KLD gradient, but the scaling factor depends on
both $s_1$ and $s_2$.  

This analysis can be generalized to non-Gaussian $p_2$, such as arbitrary exponential family distributions.
However, the final gradients depend on the derivatives of the sufficient statistics of $p_2$, which produces
a less interpretable result once taken expectation over $q_1$.

\subsubsection{Fitting mixtures distributions to intractable posteriors}\label{sec:add_exp}
For the posteriors induced by the toy distributions, 
we can optimize Gaussian mixture distributions with 10 Gaussian components 
to minimize the FD, following the biased gradient computed through
unparametrized samples drawn from the Gaussian mixture model. 
If instead we want to optimize w.r.t\ KLD, then one needs to reparametrize the discrete latent variable. 
We initialized the Gaussian components so that the means are random samples from
 the prior and the standard deviations are 1.0. The mixing proportions were
 initialized equal. All parameters are optimized using Adam with step size 0.001.
10 samples from the mixture was used to 
 approximate the FD at each iteration, for a total of 5\,000 iterations.

\begin{figure}[ht]
    \centering
    \includegraphics[width=0.8\textwidth]{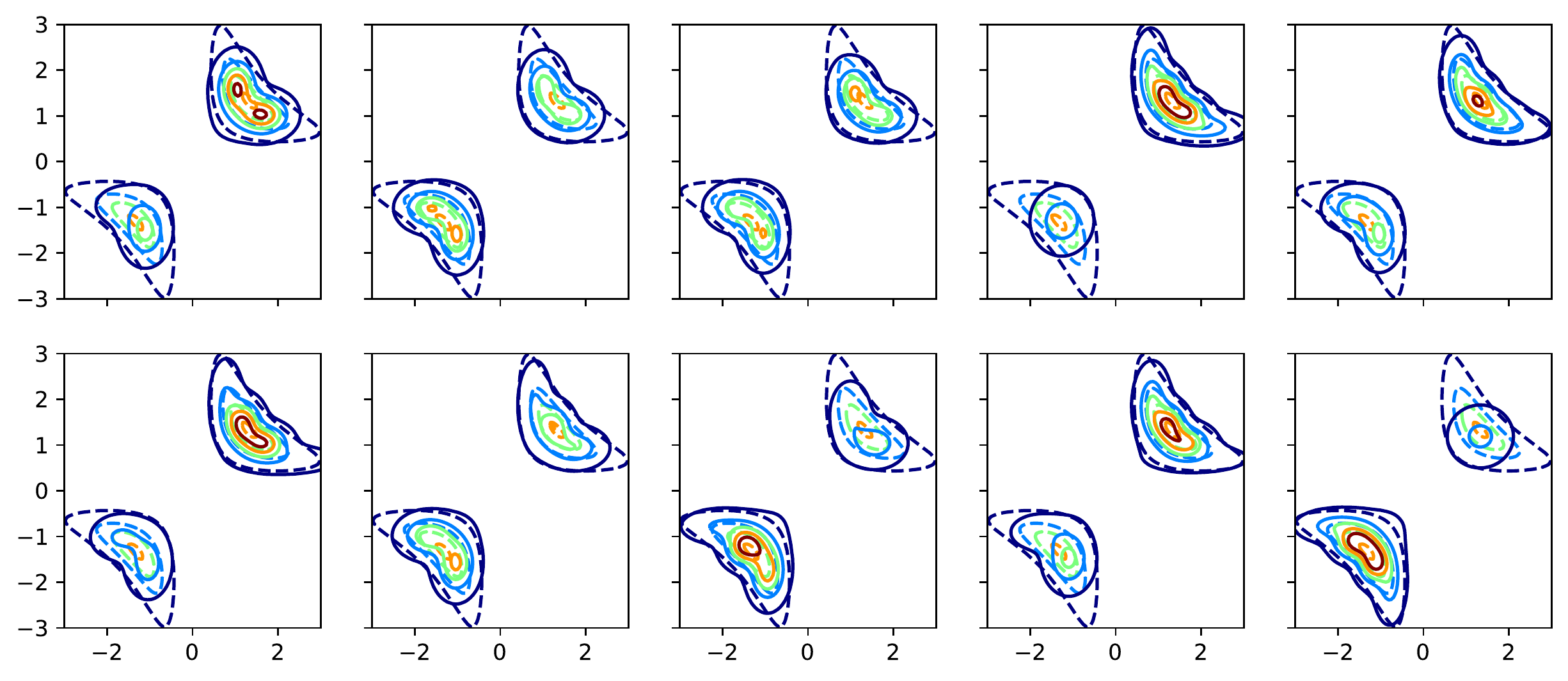}
    \includegraphics[width=0.8\textwidth]{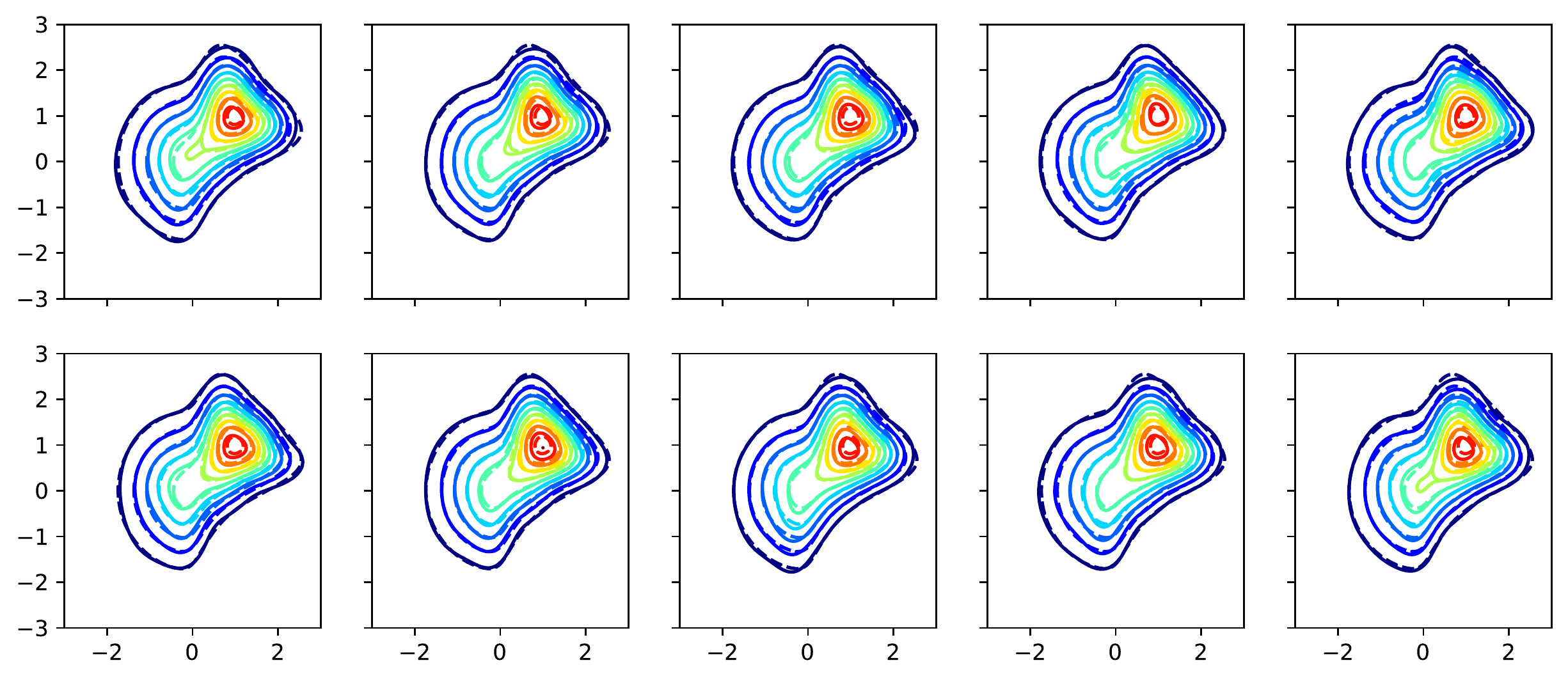}
    \caption{Fitting a Gaussian mixture distribution to posteriors induced by 
    the likelihoods \eqref{eq:toy_liks}. Each panel is a 
    run with a different initialization. Dashed line is the exact
    posterior approximated using histograms, and the solid
    lines are the model fit.}
    \label{fig:fit_gmm}
\end{figure}

To test for robustness and reliability of this method, we ran
the algorithm 10 times with different initializations.  The results 
are shown in \cref{fig:fit_gmm}.
The posterior induced by $p_\text{I}$ in \eqref{eq:toy_liks}, are bimodal with disjoint supports, 
so the fit can have arbitrary weighting between them, an issue known 
for many score-based methods. 
The fit for the posterior induced by $p_\text{II}$  was much more stable across different runs. 

\subsubsection{Fitting Laplace posterior with unparametrized samples}\label{sec:laplace}

The Lapace distribution $\text{Lap}(x;m,s)$ has a density function
$$
    \text{Lap}(x) \propto \exp\left(-\frac{|x-m|}{s}\right),
$$
and its score function is independent of the mean $m$ almost surely
$$
    \frac{\partial}{\partial x}\log \text{Lap}(x) = 
    \begin{cases}
        +\frac{1}{s} & x<m,\\
        -\frac{1}{s} & x>m.
    \end{cases}
$$
and the biased gradient will be zero for $m$. Using the unbiased gradient becomes crucial
for learning $m$.

\section{Experiment details}\label{sec:exp_details}

\subsection{Details and extended results on synthetic datasets}\label{sec:toy_exp_details}

As discussed in \cref{sec:related}, the objectives $M_2$ and $M_3$ adapted from previous work yield a biased gradient 
when the $\theta$-dependent exact posterior is replaced by a variational approximation.
To address this issue, \citet{bao2020bi} proposed a bi-level optimization technique to address the issue. 
Briefly, given one minibatch of data, before each $\vtheta$ update (learning), 
the variational parameters $\vphi$ are updated with $J$ ordinary gradient steps
followed by $K$ $\theta$-parametrized gradient steps. For $K>0$, this effectively makes $\vphi$ a function of $\vtheta$. 
Differentiating the resulting objective w.r.t.\ $\vtheta$ then gives a less biased gradient.
However, \citet{bao2020bi} set $K=$ for their large-scale experiments, showing that this 
may not be necessary. Here, we empirically test the effect of $K$ on simpler datasets and neural
architectures for interpretabiltiy.

\begin{figure}
    \centering
    \includegraphics[width=\textwidth]{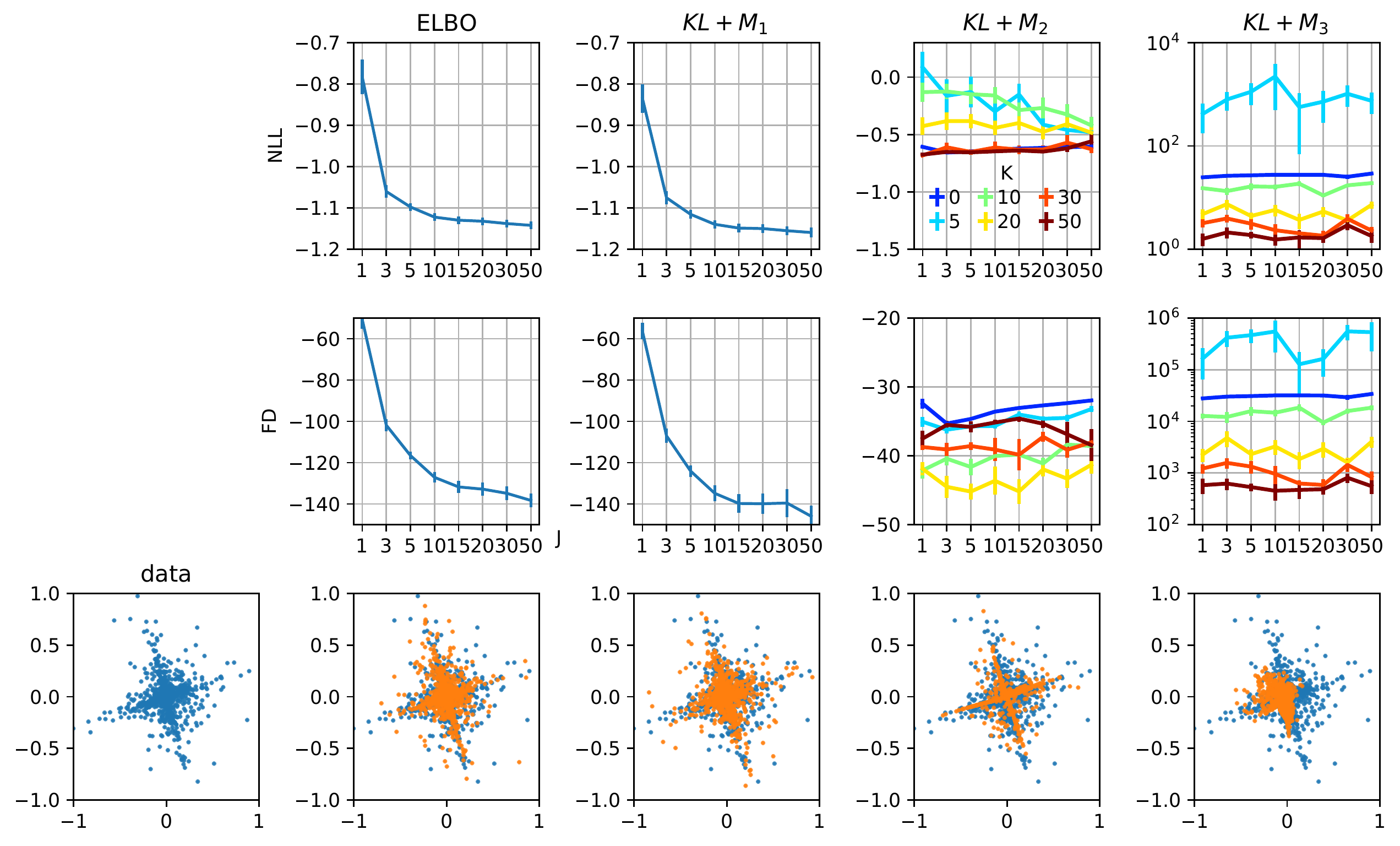}
    \caption{Test negative log-likelihood (top row) and Fisher divergence (middle row) of models trained on the star dataset. 
    Note the different scales on the vertical axes. 
    Bottom, generated and real samples of the 
    model with the highest test log-likelihood. }
    \label{fig:bilevel}
\end{figure}

\begin{figure}
    \centering
    \includegraphics[width=0.8\textwidth]{figs/banana_bilevel.pdf}\\
    \includegraphics[width=0.8\textwidth]{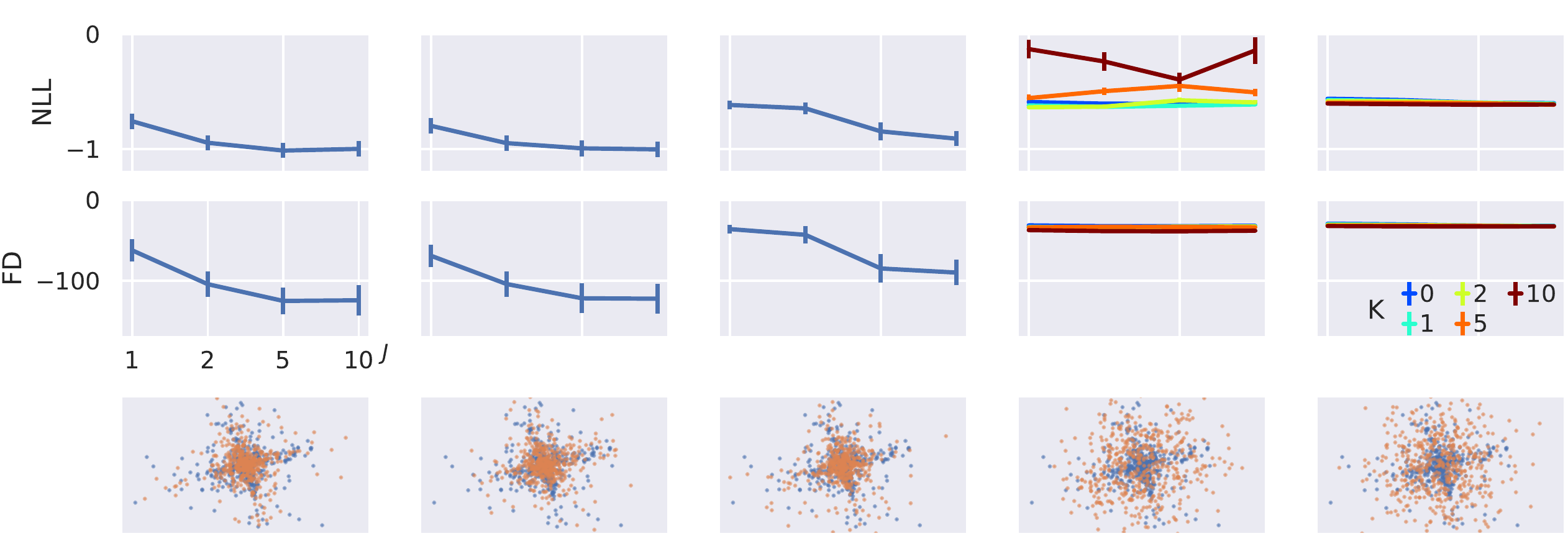}\\
    \includegraphics[width=0.8\textwidth]{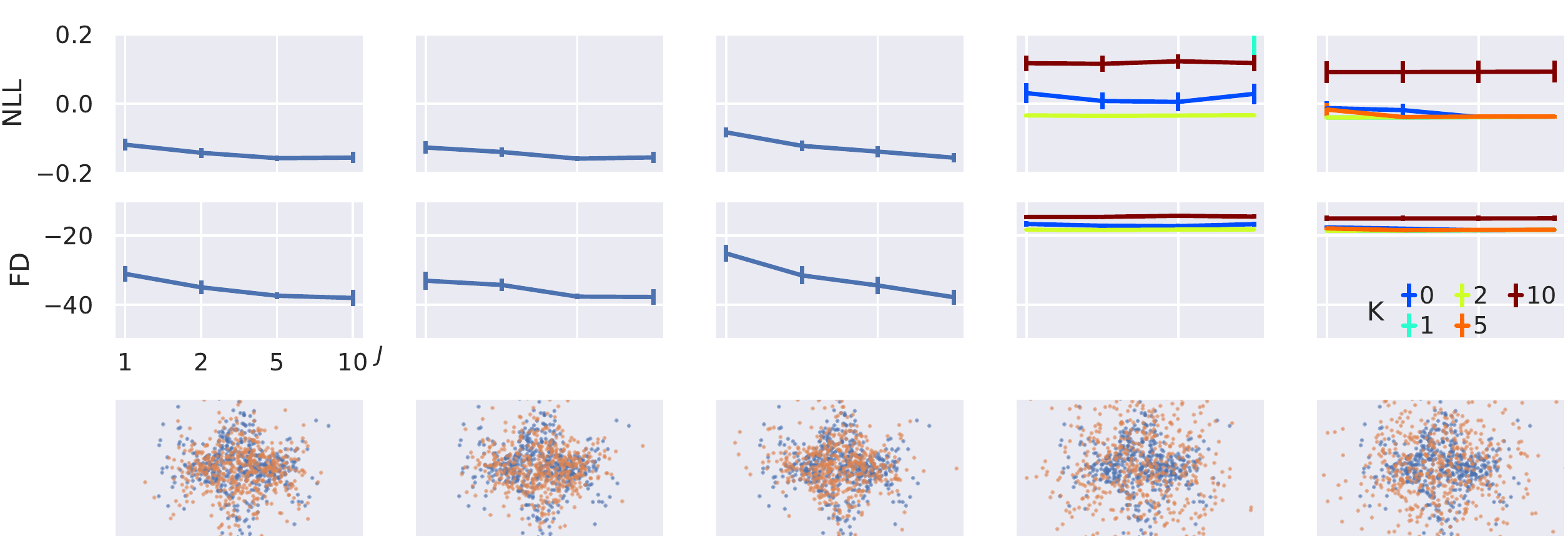}\\
    \includegraphics[width=0.8\textwidth]{figs/funnel_bilevel.pdf}\\
    \includegraphics[width=0.8\textwidth]{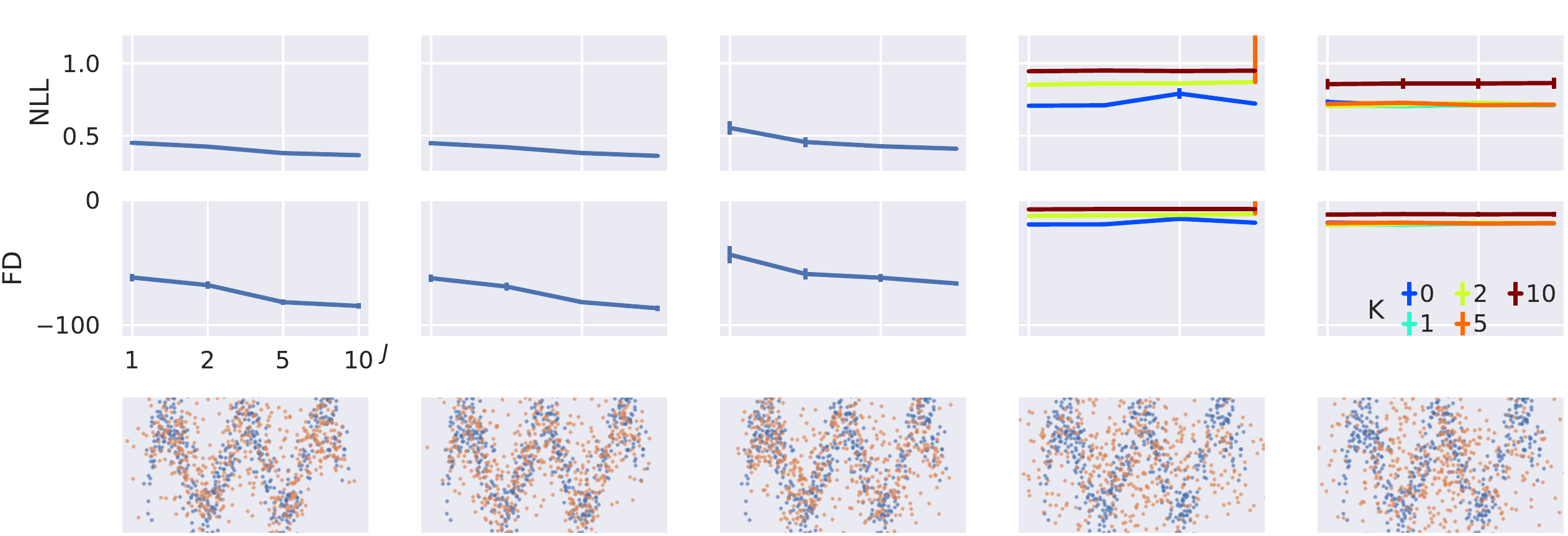}\\
    \caption{Same as \cref{fig:toy} but for all five synthetic datasets.}
    \label{fig:detailed_toy}
\end{figure}

In the main text, we show the results of training with various objectives on two datasets in \cref{fig:toy}.
We repeated the same experiment on three additional datasets, and all results are collectively 
shown in \cref{fig:detailed_toy}
In those experiments, the we used 30 neurons for each of the two hidden layers in the VAE model.
We show in \cref{fig:bilevel} the results on one dataset 
when using a larger network (two layers of 100 hidden units), and larger values of $J$ and $K$
and longer training epochs (1\,000). Here, we see that optimizing by $ELBO$ and $\KL+M_1$ produced very 
similar results with the latter being better. 
In addition, for $M_2$ and $M_3$, 
the number of bi-level updates $K$ had a substantial effect on NLL and FD. Nonetheless, 
the trained models were still much worse than ELBO and $\KL+M_1$. Thus, we reproduced the 
benefits of bi-level optimization, but these objectives still fail to fit the VAE model to this simple data set.

\subsection{Extended results on benchmark datasets}\label{sec:var_exp_details}

The latent space is $\sR^{100}$, and the images are resize to $32\times32$ (zero padding for MNIST and Fashion).
The function $\vg_\theta$ is a
neural network with DCGAN architecture and ReLU nonlinearity 
\footnote{Code taken from \url{pytorch.org/tutorials/beginner/dcgan_faces_tutorial.html}}.
The training batch size is 100. We used Adam with a step size of 0.0001 for variational methods, trained
 for 1000 epochs.
 We added a small isotropic Gaussian noise with sd 0.1 to the images. This is done for two reasons: 
 1) training with clean images or smaller std produced visible traits of overfitting on the test metrics;
 2) noise stabilizes training as $\gamma\to0$;

For each variational algorithm, we draw $S=5$  samples from $q$ (reparametrized for KLD inference and 
unparametrized for FD inference) to approximate the expectations in $M_1$ to $M_3$.
The encoder is updated for $K=5$ consecutive iterations on independent  
minibatches for every decoder update. We found 
this training procedure helped stabilize the training trajectories produced by all methods, 
including ELBO.

To evaluate the model on the test data, we computed 
\begin{itemize}
    \item reconstruction MSE: encode a noisy image, decode from the mean of $q$, and then compute the 
            mean squared error averaged over test input and the dimensionality of $\vx$. 
    \item the aggregate posterior MMD: draw one sample from $q$ for each test image, then compute 
            the MMD with samples from $p(\vz)$. We used a cubic kernel.
    \item posterior FD: we approximate $\FD[q(\vz|\vx)\|\ptheta(\vz|\vx)]$ by 5 posterior samples, then 
    average over all test $\vx$.
    \item negative ELBO: the reconstruction error is estimated by 5 samples from $q$, 
            and the KL penalty is closed-form.
\end{itemize}

\paragraph{Distribution of posterior standard deviation}

\begin{figure}
    \centering
    \includegraphics[width=\textwidth]{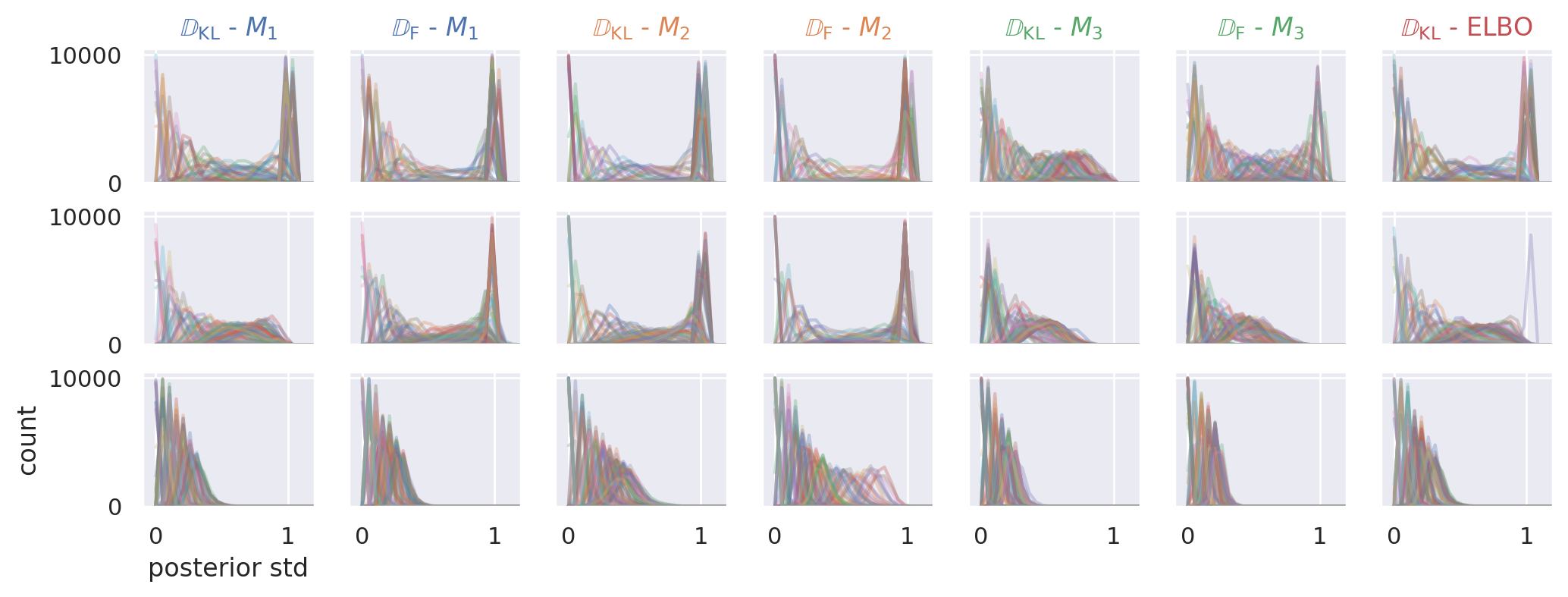}
    \caption{Histogram of posterior variance of each dimension (colored pale lines) and averaged
    over 100 dimension (black) for model with DCGAN network architecture.}
    \label{fig:ae_sparse}
\end{figure}

To verify FD-based inference has the ability to obtain minimal latent representation implied in \cref{sec:FD_inference}, 
we show the distribution of posterior standard deviations for all methods and distributions in \cref{fig:ae_sparse}.
It is clear that the standard deviations are concentrated around 0 and 1 on the MNIST dataset, consistent with the 
analysis in \cref{sec:FD_inference}. 

\begin{figure}
    \centering
    \includegraphics[width=\textwidth]{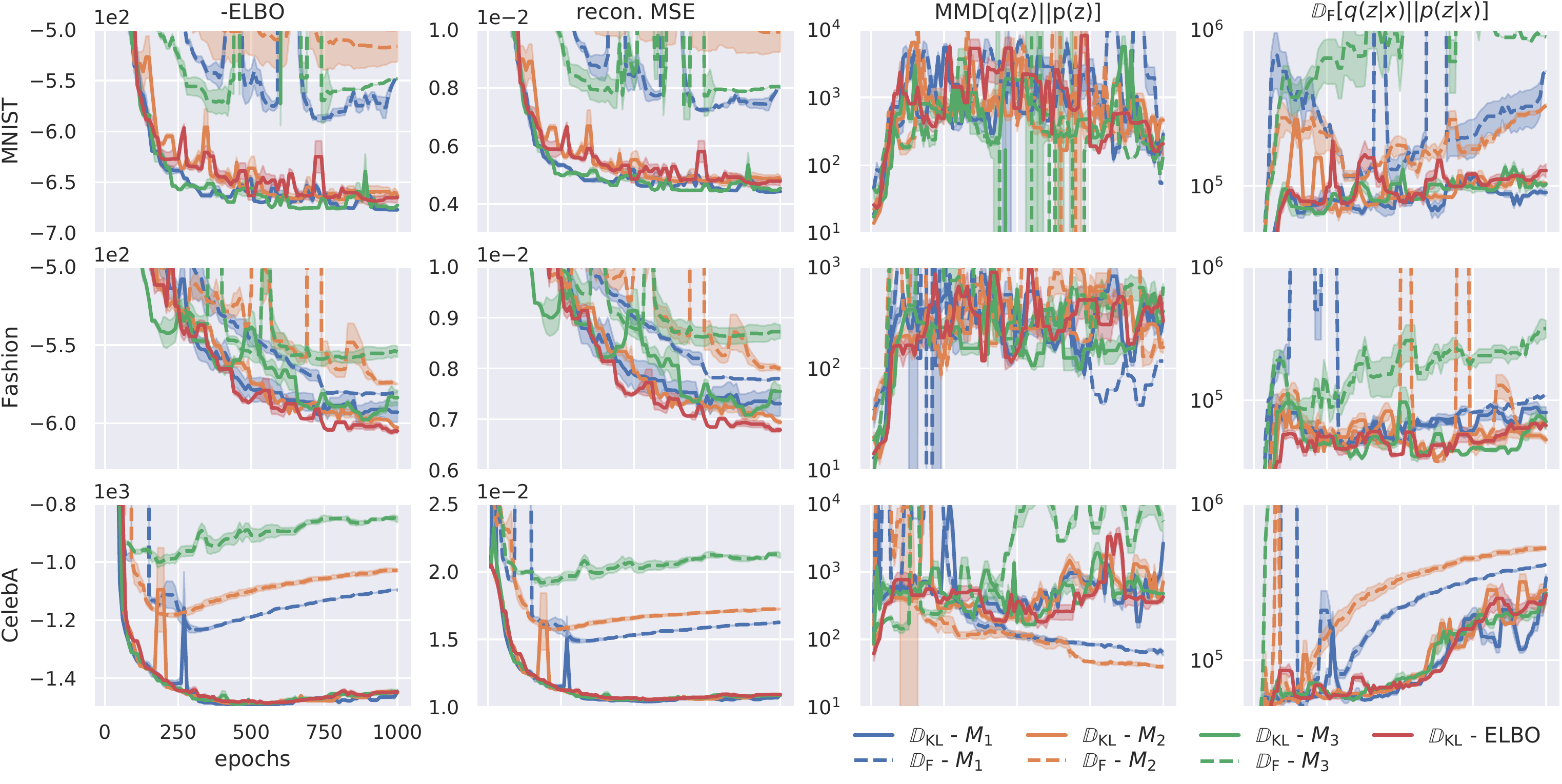}
    \caption{Same as \cref{fig:ae} but for VAE models with ResNet18 structure.}
    \label{fig:ae_resnet}
\end{figure}

\paragraph{Experiments on ResNet18}
To test whether the observations so far hold for other network architectures, we 
ran the variational SM objectives and ELBO on VAE models with ResNet18\footnote{Code taken from \url{github.com/julianstastny/VAE-ResNet18-PyTorch}.}. 
Here, we set $J=1$ and $S=1$.
Again, KL-based inference gave better metrics on all the four meausres. 
The performance metrics are shown in \cref{fig:ae_resnet}, averaged over five runs. 
Again $\sK+M_1$ performed almost identical to 
ELBO, but now $\KL+M_2$ and $\KL+M_3$ also became similar to those two. 
This was not the case for the fully connected networks on synthetic datasets or 
convnets on benchmark dataset. Therefore, the network architecture seems to allow
$M_2$- and $M_3$-based methods to improve quite substantially.

\begin{figure}
    \centering
    \includegraphics[width=\textwidth]{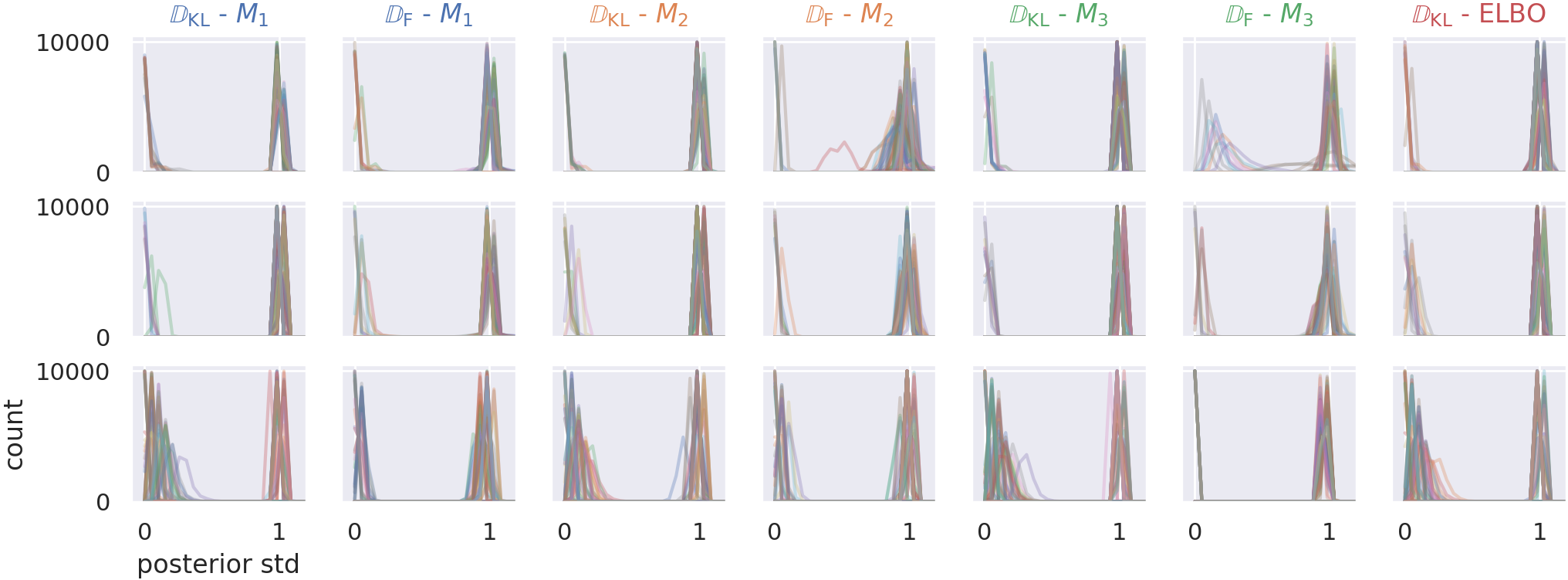}
    \caption{Same as \cref{fig:ae_sparse} but using models with ResNet architecture.}
    \label{fig:ae_sparse_resnet}
\end{figure}

On sample quality, 
shown in \cref{tab:fids_resnet}, $M_2$ gave the best qualities while $M_3$ were the worst. 
A notable difference is that Objectives with FD-based inference could 
lead to better sample quality.
Objectives with FD-based inference can lead to better sample quality than those with KLD-based inference, 
although the ELBO or posterior quality were worse \cref{fig:ae_sparse_resnet}.
The histograms of posterior sds \cref{fig:ae_sparse_resnet} become more concentrated around 0 and 1 
using ResNets than using ConvNets for all objectives, 
suggesting better capture of the data manifold by the latent codes in ResNet models.


\paragraph{Experiments on binary latent variables}

To test whether the minimal representation does not rely on a Gaussian prior,
we trained a VAE model with binary $\vz\in\{0,1\}^30$. 
We used a factorized Bernoulli $q$ whose parameters are optimized by Gumble Softmax.

\begin{figure}
    \centering
    \includegraphics[width=\textwidth]{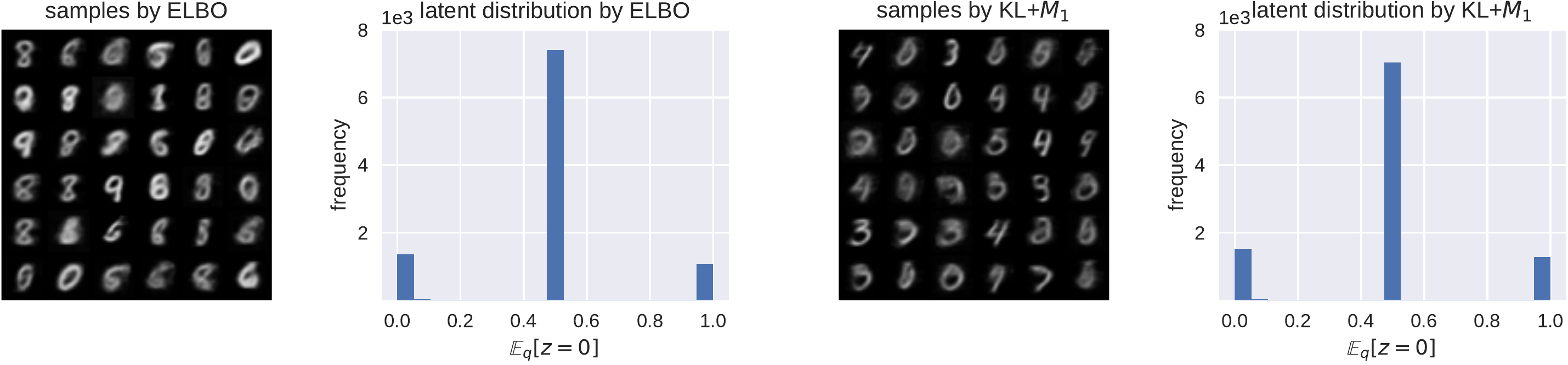}
    \caption{Results of a VAE with binary latent variables, trained by $\KL$-ELBO (left two panels) and 
    $\KL$-$M_1$ (right two panels). For each objective, we show the generated samples and 
    the distributions of variational posterior mean conditioned on test data samples.
    }
    \label{fig:binary_sparse}
\end{figure}

If a given dataset 
requires only a few latent codes to describe, then the redundant latent space
will have their posteriors match the prior. In \cref{fig:binary_sparse}, we 
see that is the case when the VAE model is trained by ELBO or $\KL$+$M_1$. 
The histograms show the distributions of the posterior means. 
Many of the latent dimensions have mean 0.5, suggesting that these dimensions 
do not encode meaning information about the data.

\begin{table}[t!]
    \setlength{\tabcolsep}{2.5pt}
    \centering
    \label{tab:fids_resnet}
    \caption{Sample quality (KID $\times 1\,000$) for variational models trained with ResNet18 architecture. 
    Lower is better. Bold indicates the best.
    \vspace{1ex}
    }\label{tab:allexp_resnet}
    {
    \begin{tabular}{ccccccc}
        \toprule
        \multicolumn{1}{}{} & \multicolumn{2}{c}{\textbf{MNIST}} & \multicolumn{2}{c}{\textbf{Fashion}} & \multicolumn{2}{c}{\textbf{CelebA}}\\
        \multicolumn{1}{c}{\textbf{Method}}              & FID & KID  & FID & KID & FID & KID\\
        \cmidrule(r){1-7}
        \multicolumn{1}{c}{{Baseline}} & \multicolumn{2}{c}{} & \multicolumn{2}{c}{} & \multicolumn{2}{c}{}\\
        \multicolumn{1}{c}{VAE}  &  5.58 & 96.8 & 4.89 & 81.1 & 55.4 & 56.9\\
        \cmidrule(r){1-7}
        \multicolumn{1}{c}{{Variational}} & \multicolumn{2}{c}{} & \multicolumn{2}{c}{} & \multicolumn{2}{c}{}\\
        \multicolumn{1}{c}{$\sK$+$M_1$} & 5.24  & 90.2 & 4.63 & 73.5 & 54.6 & 56.1\\
        \multicolumn{1}{c}{$\sF$+$M_1$} & 2.93 & 54.7 & 4.86 & 79.4 & 59.2 & 58.6\\
        \multicolumn{1}{c}{$\sK$+$M_2$} & 5.81 & 99.8 & 4.77 & 76.0 & \textbf{54.3} & \textbf{55.1}\\
        \multicolumn{1}{c}{$\sF$+$M_2$} & \textbf{2.88} & \textbf{44.3} & \textbf{3.65} & \textbf{52.2} & 60.4 & 60.0\\
        \multicolumn{1}{c}{$\sK$+$M_3$} & 12.6 & 264 & 10.7  & 205  & 208 & 226\\
        \multicolumn{1}{c}{$\sF$+$M_3$} & 8.70 & 177 & 28.6  & 637  & 203 & 196\\
        \bottomrule
    \end{tabular}
    }
\end{table}

\subsection{Generated samples}
\begin{figure}
    \centering
    \includegraphics[width=\textwidth]{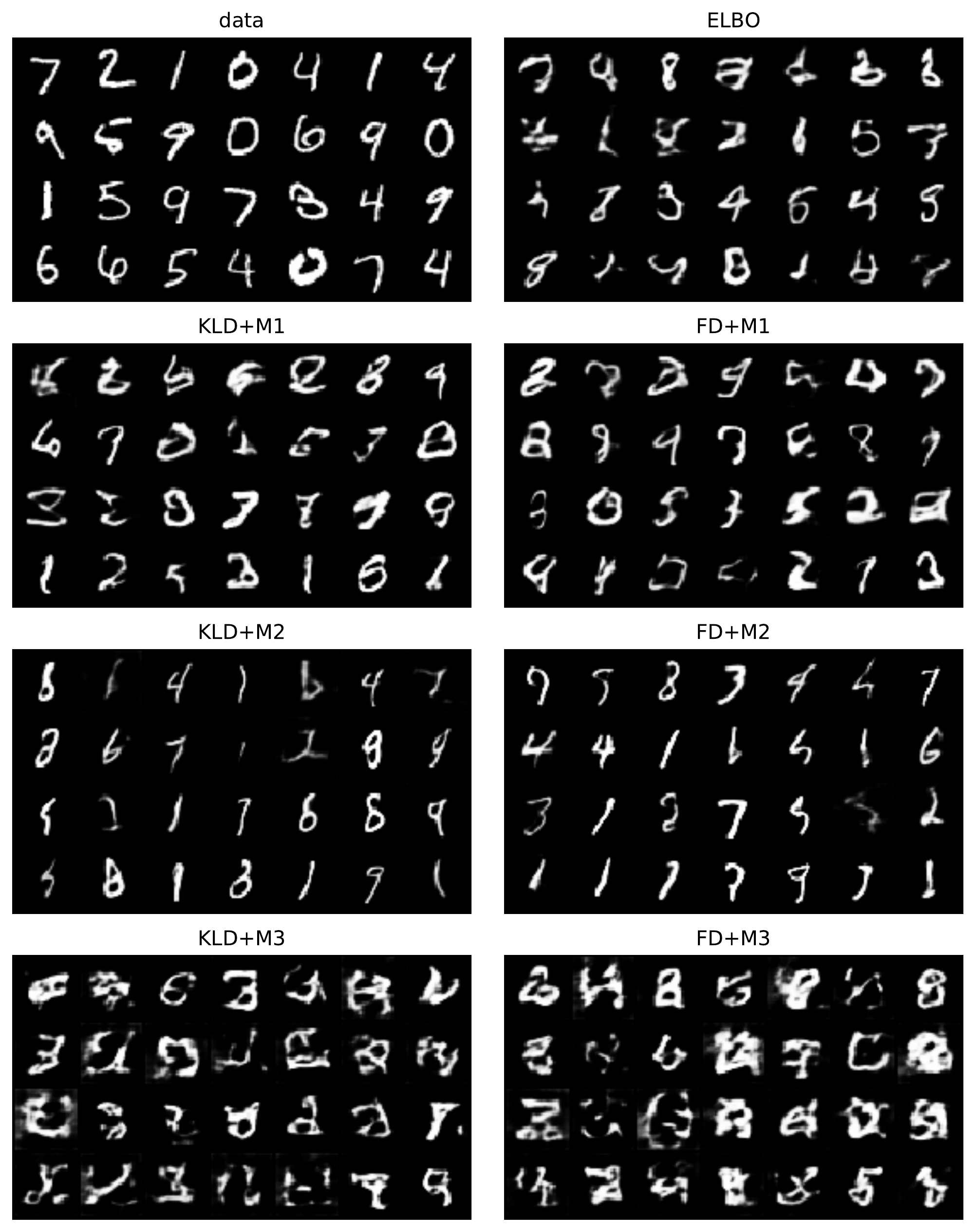}
    \caption{MNIST samples from ConvNet.}
    \label{fig:conv_mnist}
\end{figure}

\begin{figure}
    \centering
    \includegraphics[width=\textwidth]{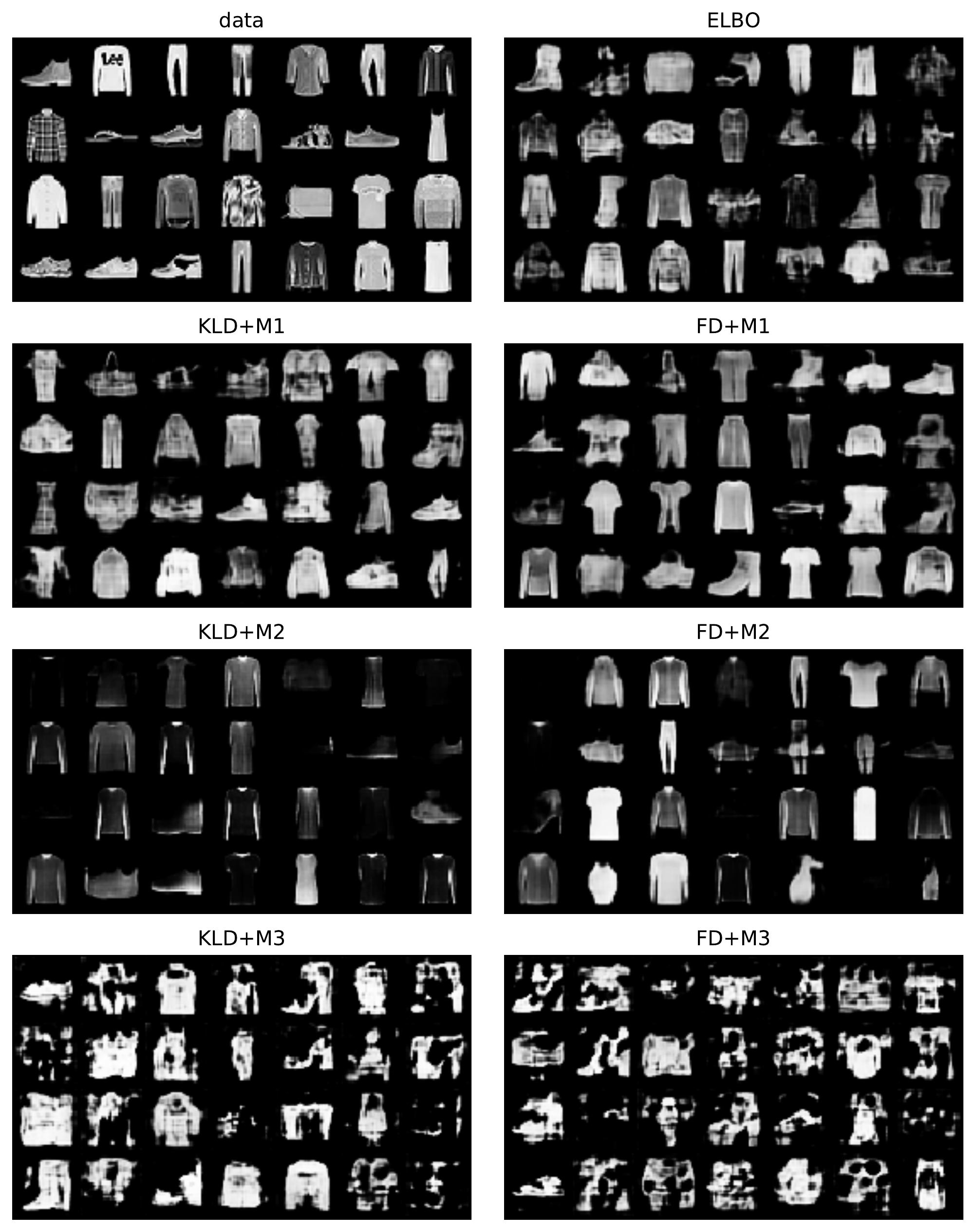}
    \caption{Fashion samples from ConvNet.}
    \label{fig:conv_fashion}
\end{figure}

\begin{figure}
    \centering
    \includegraphics[width=\textwidth]{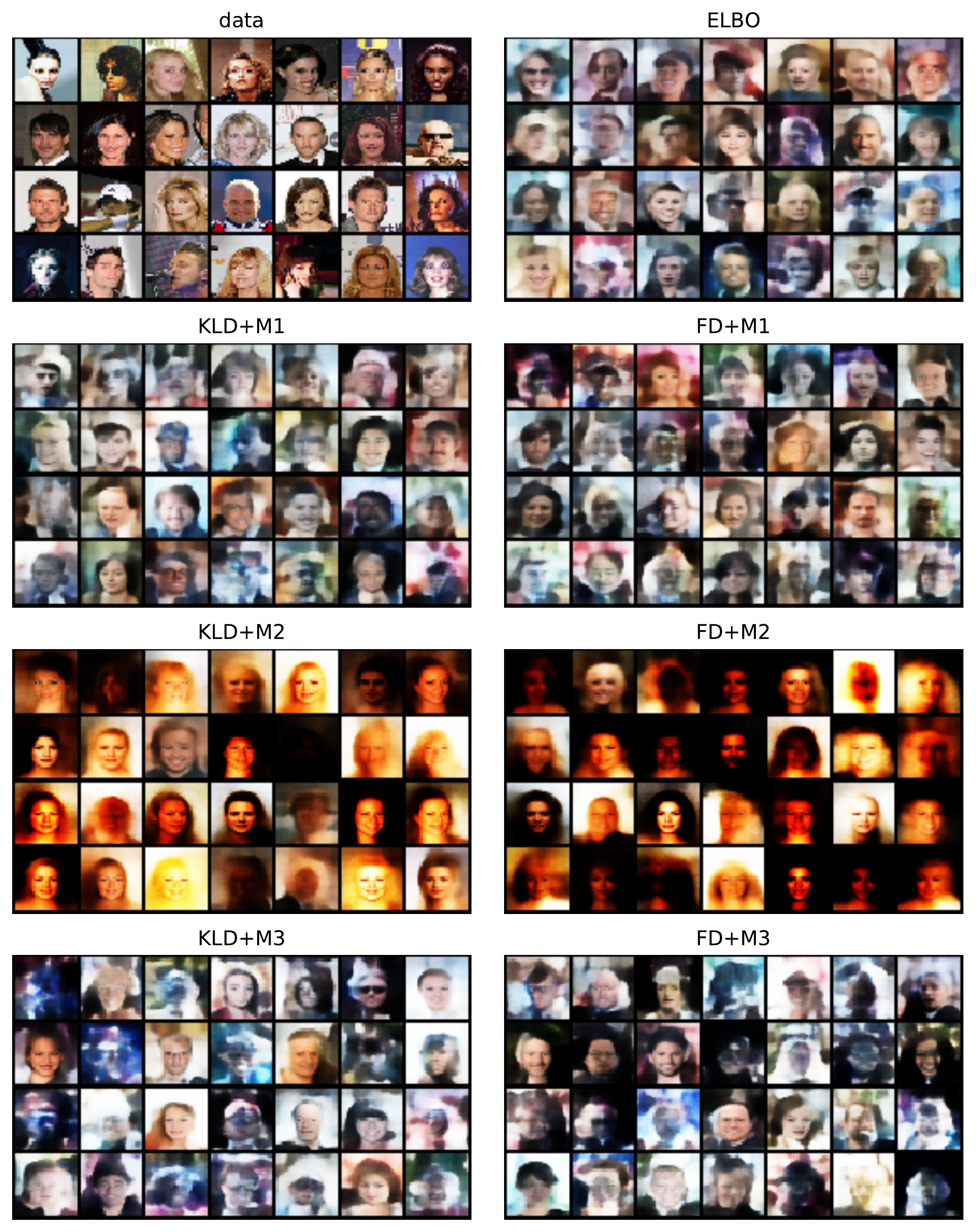}
    \caption{CelebA samples from ConvNet.}
    \label{fig:conv_celeb}
\end{figure}

\begin{figure}
    \centering
    \includegraphics[width=\textwidth]{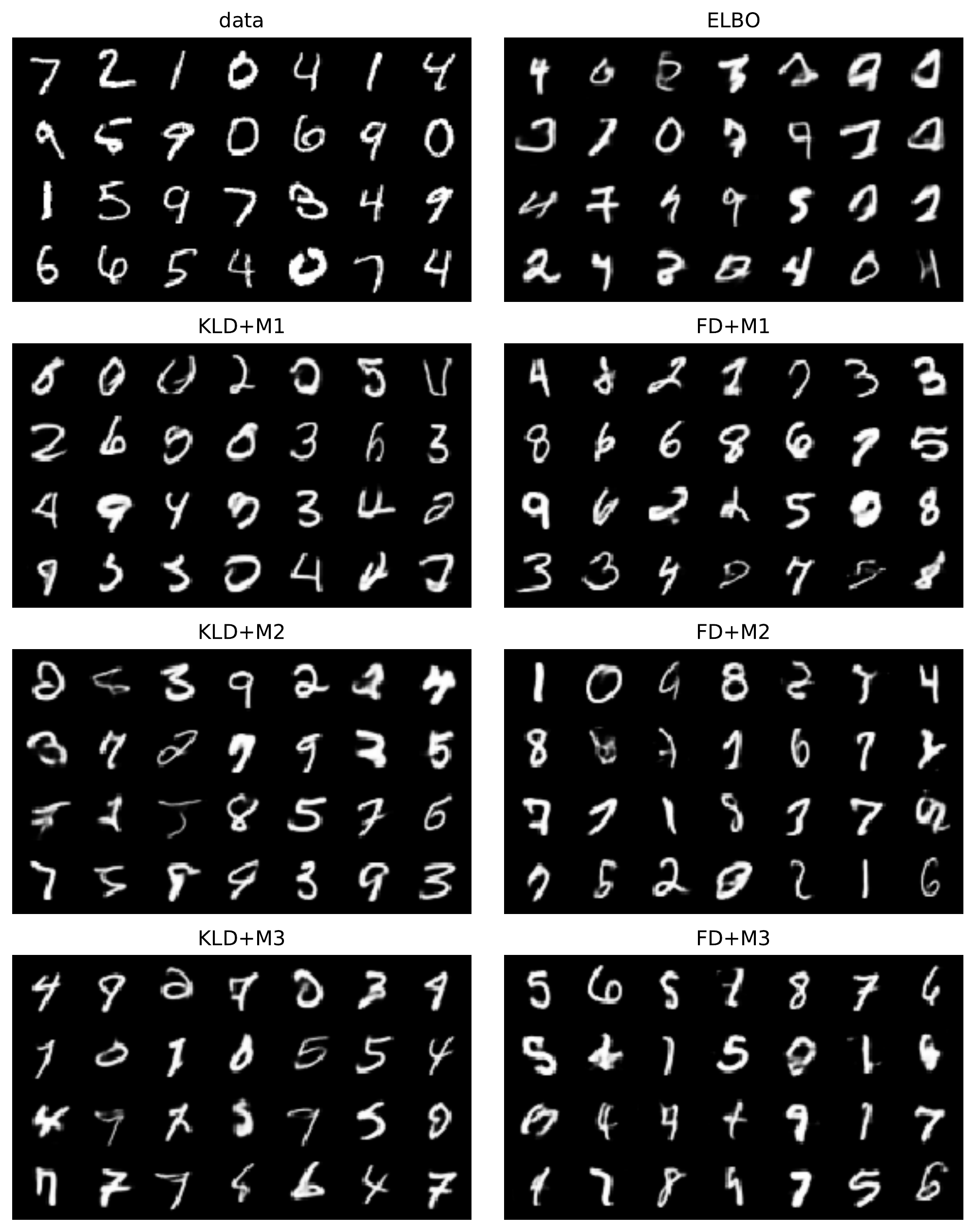}
    \caption{MNIST samples from ResNet.}
    \label{fig:resnet_mnist}
\end{figure}

\begin{figure}
    \centering
    \includegraphics[width=\textwidth]{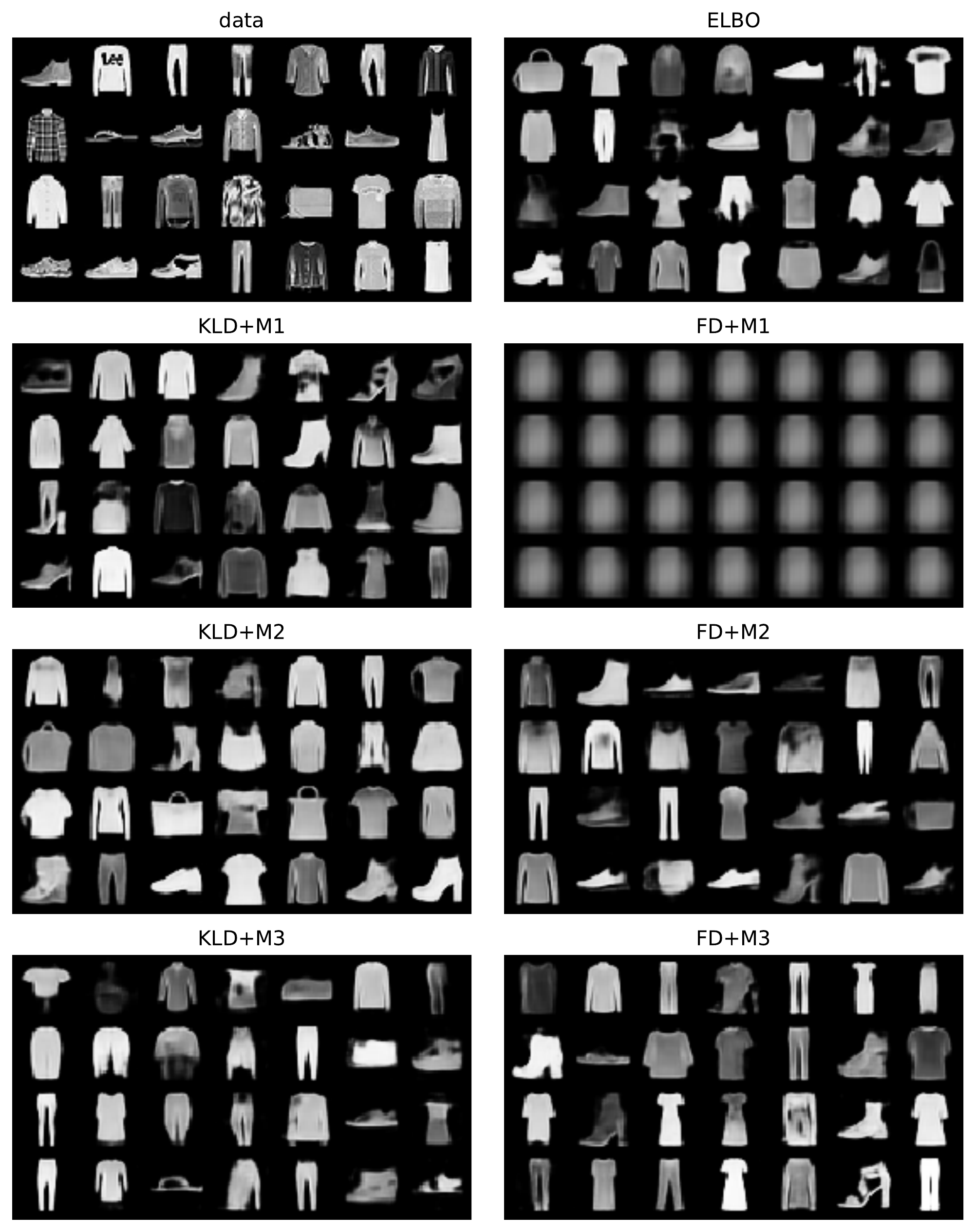}
    \caption{Fashion samples from ResNet.}
    \label{fig:resnet_fashion}
\end{figure}

\begin{figure}
    \centering
    \includegraphics[width=\textwidth]{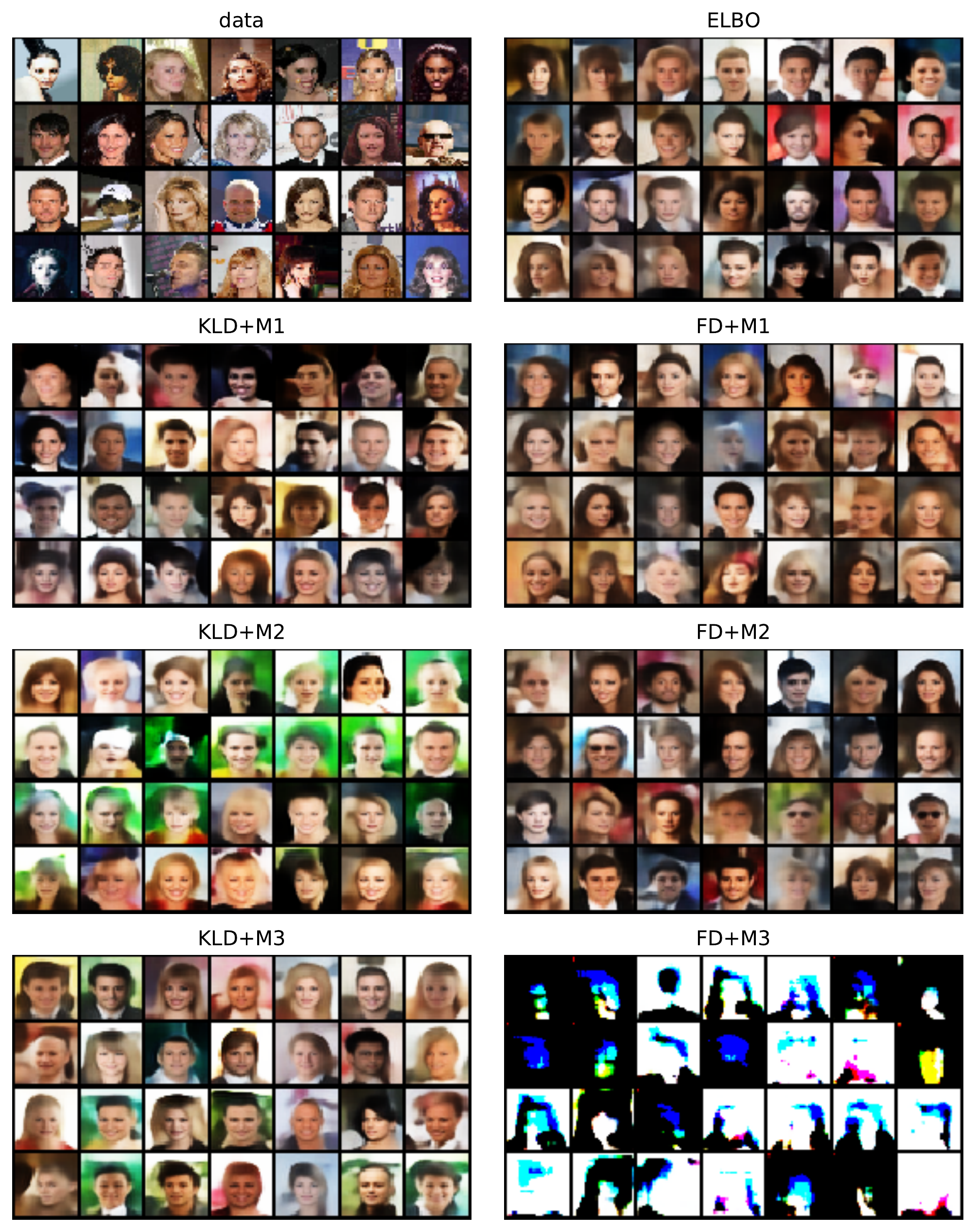}
    \caption{CelebA samples from ResNet.}
    \label{fig:resnet_celeb}
\end{figure}

The generated samples in \cref{fig:conv_mnist,fig:conv_fashion,fig:conv_celeb} are drawn from 
models with ConvNets, and 
\cref{fig:resnet_mnist,fig:resnet_fashion,fig:resnet_celeb} with ResNets.

 

\end{document}

%% file: math_commands.tex
\usepackage{amsmath,amsfonts,bm,amssymb}

\def\1{\bm{1}}

\def\vmu{{{\mu}}}

\def\vgamma{{\bm{\gamma}}}

\def\vtheta{{{\theta}}}
\def\vphi{{{\phi}}}

\def\vepsilon{\bm{\epsilon}}

\def\vd{{{d}}}

\def\vf{{{f}}}
\def\vg{{{g}}}

\def\vs{{{s}}}

\def\vx{{{x}}}

\def\vz{{{z}}}

\def\mG{{{G}}}

\def\mI{{{I}}}

\def\mLambda{{{\Lambda}}}
\def\mSigma{{{\Sigma}}}

\DeclareMathAlphabet{\mathsfit}{\encodingdefault}{\sfdefault}{m}{sl}
\SetMathAlphabet{\mathsfit}{bold}{\encodingdefault}{\sfdefault}{bx}{n}

\def\gD{{\mathcal{D}}}

\def\gF{{\mathcal{F}}}

\def\gL{{\mathcal{L}}}

\def\gN{{\mathcal{N}}}

\def\gQ{{\mathcal{Q}}}
\def\gR{{\mathcal{R}}}

\def\sC{{\mathbb{C}}}
\def\sD{{\mathbb{D}}}
\def\sF{{\mathbb{F}}}

\def\sK{{\mathbb{K}}}

\def\sR{{\mathbb{R}}}

\def\sV{{\mathbb{V}}}

\DeclareMathOperator{\E}{\mathbb{E}}
\newcommand{\Eqpi}{\E_{q\pi}\!}
\newcommand{\Eq}{\E_q\!}

\DeclareMathOperator{\Img}{\text{\Img}}
\DeclareMathOperator{\Rot}{\text{\Rot}}

\newcommand{\R}{\mathbb{R}}

\newcommand{\KL}{\sD_{\mathrm{KL}}}
\newcommand{\FD}{\sD_{\mathrm{F}}}

\newcommand{\nx}{{\nabla_{\vx}}}
\newcommand{\nz}{{\nabla_{\vz}}}

\DeclareMathOperator{\Tr}{Tr}

\newcommand{\ud}{\textnormal{d}}